\documentclass[fleqn,10pt]{wlscirep}
\usepackage[utf8]{inputenc}
\usepackage[T1]{fontenc}
\usepackage{dirtree}

\title{A dataset of medication images with instance segmentation masks for preventing adverse drug events}

\author[1,*]{Wai Ip Chu}
\author[2]{Shashi Hirani}
\author[1,3]{Giacomo Tarroni}
\author[1,+]{Ling Li}
\affil[1]{City St George's, University of London, School of Science \& Technology, London, EC1V 0HB, United Kingdom}
\affil[2]{City St George's, University of London, School of Health \& Medical Sciences, London, EC1V 0HB, United Kingdom}
\affil[3]{Imperial College London, Dept of Computer Science, London, SW7 2AZ, United Kingdom}

\affil[*]{William.Chu@citystgeorges.ac.uk}
\affil[+]{Caroline.Li@citystgeorges.ac.uk}


\begin{abstract}
Medication errors and adverse drug events (ADEs) pose significant risks to patient safety, often arising from difficulties in reliably identifying pharmaceuticals in real-world settings. AI-based pill recognition models offer a promising solution, but the lack of comprehensive datasets hinders their development. Existing pill image datasets rarely capture real-world complexities such as overlapping pills, varied lighting, and occlusions. MEDISEG addresses this gap by providing instance segmentation annotations for 32 distinct pill types across 8,262 images, encompassing diverse conditions from individual pill images to cluttered dosette boxes. We trained YOLOv8 and YOLOv9 on MEDISEG to demonstrate their usability, achieving mAP@50 of 99.5\% on the 3-Pills subset and 80.1\% on the 32-Pills subset. We further evaluate MEDISEG under a few-shot detection protocol, demonstrating that base training on MEDISEG significantly improves recognition of unseen pill classes in occluded multi-pill scenarios compared to existing datasets. These results highlight the dataset’s ability not only to support robust supervised training but also to promote transferable representations under limited supervision, making it a valuable resource for developing and benchmarking AI-driven systems for medication safety. 
\end{abstract}
\begin{document}

\flushbottom
\maketitle
\thispagestyle{empty}

\section*{Background \& Summary}

Adverse events of medical treatments (AEMT) and adverse drug events (ADE) are critical challenges in healthcare, representing significant risks to patient safety. AEMT encompasses any harmful outcome resulting from medical interventions, ranging from surgical complications and diagnostic errors to issues with medical devices and medications \cite{skelly:ae:2023}. In contrast, ADE specifically refers to the harm caused by medications, including errors such as incorrect dosage and adverse drug reactions (ADR) that occur even when medications are administered correctly (e.g., incorrect dosages or combinations of interacting drugs)\cite{nebeker:ade:2004}. 

\begin{figure}[ht]
\centering
\includegraphics[width=\linewidth]{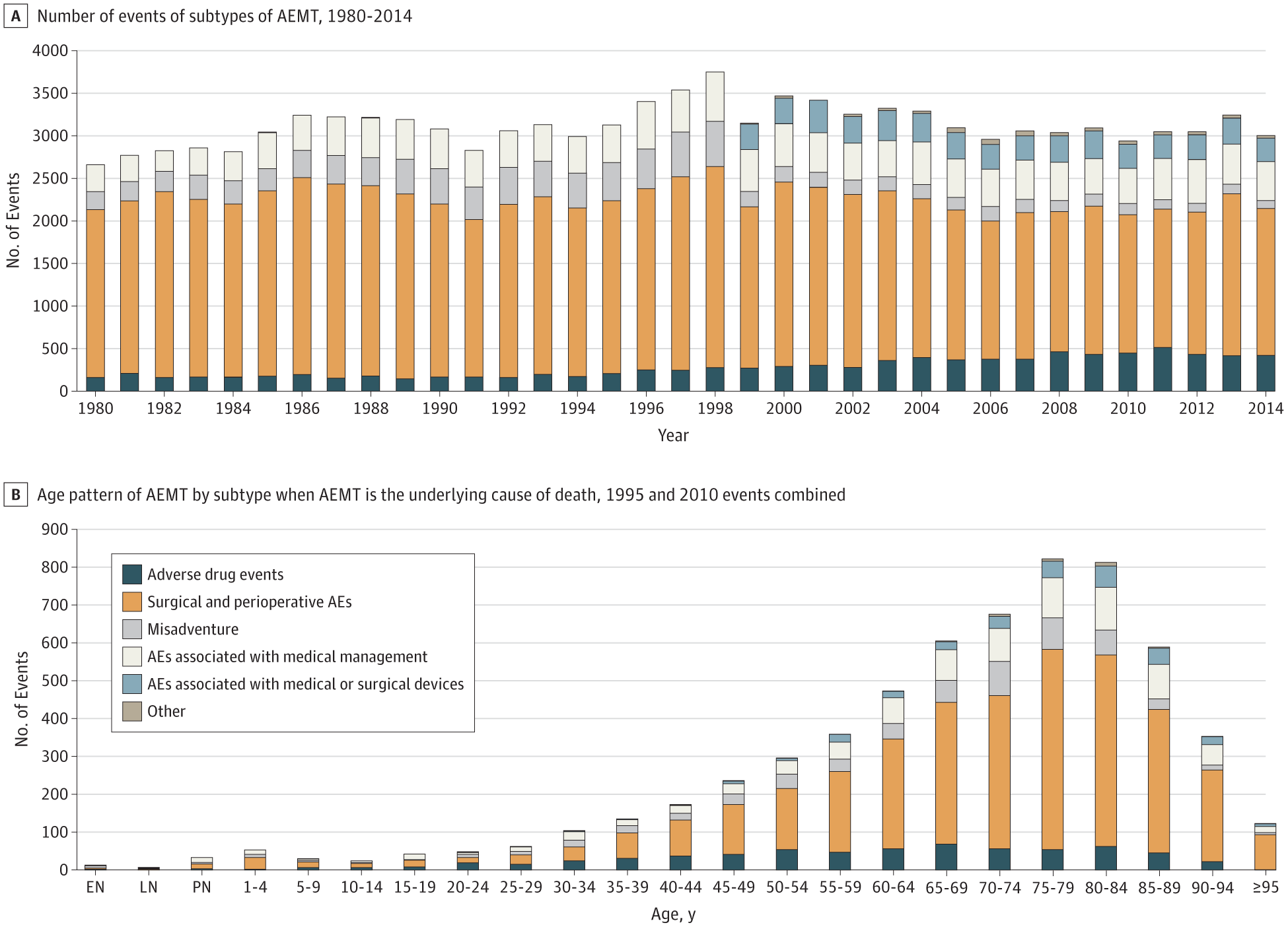}
\caption{(A) Trends in AEMT events from 1980 to 2014 and (B) the distribution of these events across age groups \cite{sunshine:aemt:2019}.}
\label{fig:AEMT_subtypes}
\end{figure}

A study by Sunshine\cite{sunshine:aemt:2019} provides a comprehensive overview of mortality associated with AEMT in the United States from 1980 to 2014. Figure \ref{fig:AEMT_subtypes}A demonstrates that, although the overall number of AEMT events has stayed relatively constant throughout the years, incidents associated with ADE have proportionally exhibited a consistent rise. Specifically, an average of 8.9\% of fatalities for which AEMT was the primary cause were ascribed to ADE over the period from 1980 to 2014. Figure \ref{fig:AEMT_subtypes}B further reveals that advancing age is associated with an increased risk of ADEs, likely due to polypharmacy, age-related physiological changes, and a higher prevalence of chronic conditions. In contrast, younger individuals face disproportionately higher mortality rates from these events, potentially due to misadministration or accidental exposure. Given these trends, AI-driven solutions for pill recognition could help mitigate ADE risks, particularly in vulnerable populations such as the elderly and individuals with complex medication regimens.

\subsection*{Existing Pill Datasets and Their Limitations}

Numerous datasets have been developed to support the advancement of pill recognition systems. Notable examples include the dataset from Lee et al.\cite{lee:Pill-ID:2012}, the National Institutes of Health (NIH) Pillbox dataset\cite{yaniv:nlm:2016}, the CURE dataset\cite{ling:few-shot:2020}, the dataset from Wong et al.\cite{wong:finegrain:2017}, and the dataset from Tan et al.\cite{tan:comparison:2021}. Table \ref{tab:existing_dataset_comparison} compares these five datasets. For instance, the NIH Pillbox dataset stands out with 4,392 high-quality reference images and 133,774 consumer-grade images, yet it lacks detailed instance segmentation labels. Likewise, although the CURE dataset contains some instance segmentation labels, their incompleteness throughout the dataset restricts its applicability to specific computer vision tasks. Tan's dataset, with 5,131 images, employs bounding box annotations on every image, making it valuable for object detection.

\begin{figure}[ht]
\centering
\includegraphics[width=\linewidth]{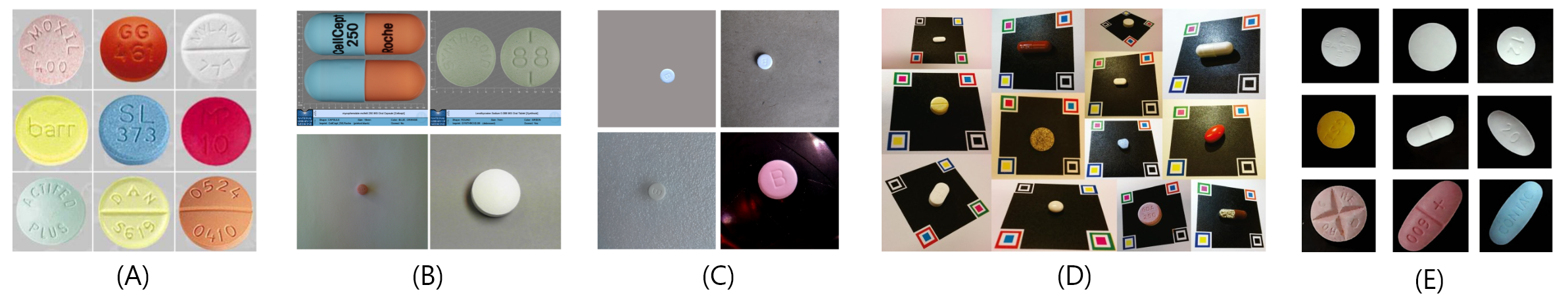}
\caption{Examples of images taken from (A) the dataset by Lee et al.\cite{lee:Pill-ID:2012}, (B) the NIH Pillbox dataset\cite{yaniv:nlm:2016}, (C) the CURE dataset\cite{ling:few-shot:2020}, (D) the dataset by Wong et al.\cite{wong:finegrain:2017}, and (E) the dataset by Tan et al.\cite{tan:comparison:2021}.}
\label{fig:datasets}
\end{figure}

Figure \ref{fig:datasets} highlights examples from these datasets. Notably, of the limited publicly available datasets, the CURE dataset also includes synthetic images in its collection. Although synthetic data can help expand training sets, it often fails to capture real-world variations. A common limitation shared across all these datasets is their controlled acquisition environment. Images are typically taken under optimal lighting conditions with high-quality cameras and feature single, unobstructed pills. Such conditions do not reflect the complexity of real-world scenarios where pills are found in cluttered environments, under variable lighting, or within dosette boxes with overlapping or partially occluded pills.

The need for a dataset incorporating multi-pill images becomes particularly relevant given the increasing medication burden associated with ageing. Studies report that more than a third of adults between 75 and 85 take at least five prescription medications daily\cite{reves:medication:2023}. As the number of medications per patient rises, so does the risk of unintentional medication errors and non-adherence issues that stem not from intentional refusal but from confusion, misidentification, or accidental omission. These errors can lead to severe health consequences, particularly in older populations with complex medication regimens. AI-powered pill recognition systems, trained on datasets that capture real-world multi-pill conditions, offer a promising solution by providing automated identification and validation, thereby reducing the likelihood of errors in medication administration and improving adherence in high-risk groups.

\subsection*{Motivation for the MEDISEG dataset}

The limitations of current pill datasets, especially the emphasis on individual pill images and insufficient instance segmentation annotations, highlight the necessity for a dataset that reflects real-world intricacies. Only the NIH Pillbox and CURE datasets are publicly available datasets. The National Library of Medicine's Pillbox program was discontinued on January 29, 2021, ending the creation of the Pillbox dataset, its image repository, and related APIs. While a repository of reference images exists, these photos lack accompanying label files, making them inadequate for effective model training. Moreover, the consumer images, essential for depicting the varied and hard circumstances faced in daily situations, are no longer available. 

To address these shortcomings, we developed the MEDication Image SEGmentation (MEDISEG) dataset. MEDISEG captures naturally occurring scenarios, such as multiple pills in dosette boxes under varied lighting and environmental conditions. Each image is meticulously annotated with instance segmentation masks, providing a robust foundation for training advanced pill recognition models.

\section*{Methods}

This section provides a step-by-step overview of how the MEDISEG dataset was created, from the initial data acquisition and image capture setup to preprocessing and annotation procedures. A final comparison with existing datasets illustrates how MEDISEG addresses previously identified limitations in pill recognition data.

\subsection*{Data Acquisition and Image Capture}

All images were captured using an iPhone 12 Pro Max, which provides high-quality images that can later be preprocessed to simulate various imaging conditions as needed. Artificial lighting was manipulated in both intensity and angle during various picture sessions, creating authentic shadows, reflections, and highlights.

\begin{figure}[ht]
\centering
\includegraphics[width=\linewidth]{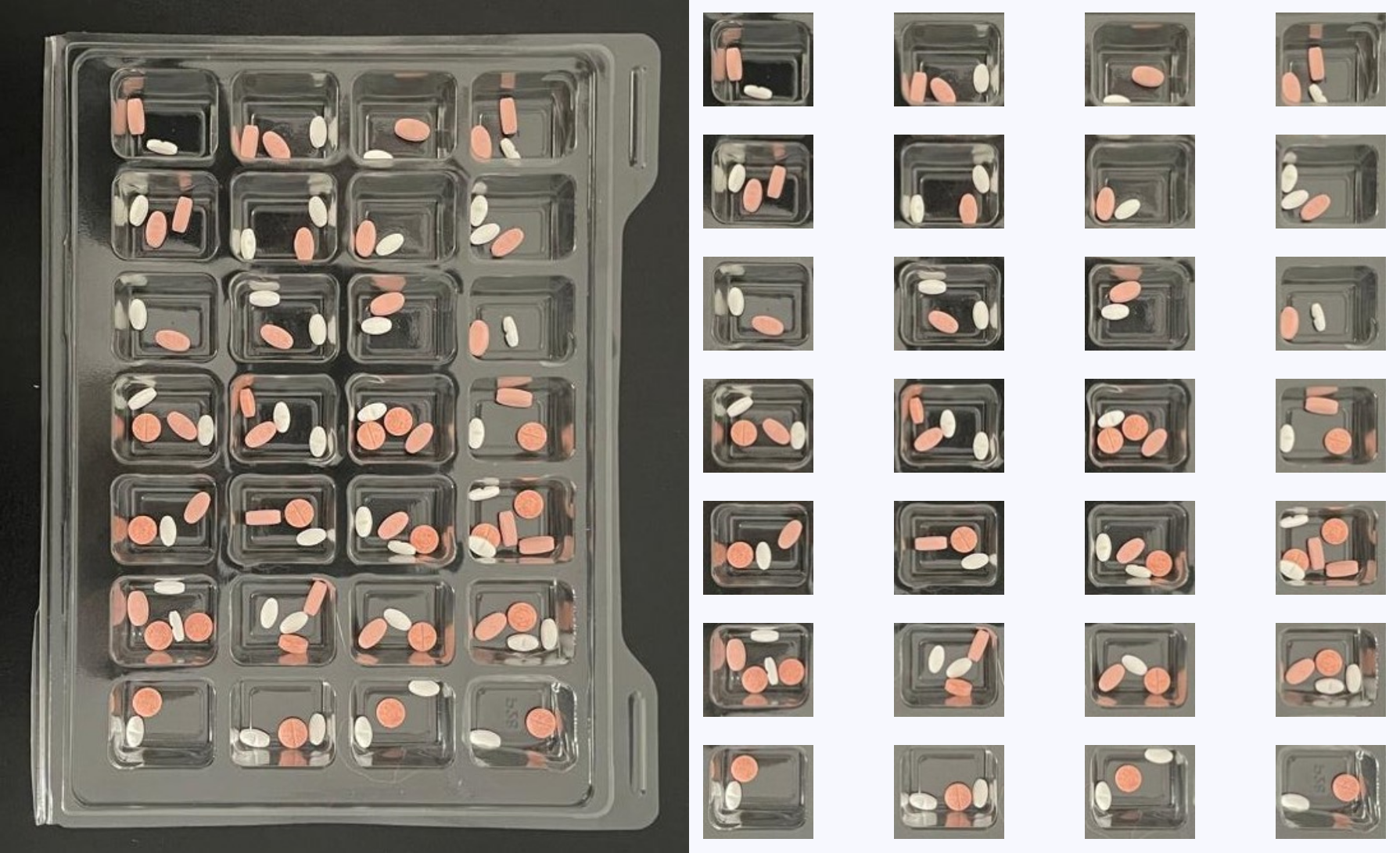}
\caption{Pills organised in a standard four-by-seven dosette box (left) and after cropping into individual images (right).}
\label{fig:cropped_images}
\end{figure}

A standard four-by-seven dosette box was used to arrange pills, as shown in Figure \ref{fig:cropped_images} (left). This configuration offers a pragmatic framework for each image while integrating real-world intricacies, including overlapping tablets and partial obstructions due to the dosette box walls.

\subsection*{Image Preprocessing and Annotation}

Following image capture, each dosette box image was cropped to isolate individual pill slots, as shown in Figure \ref{fig:cropped_images} (right), reducing background distractions and introducing a small degree of camera angle variation. The cropped images were then padded and resized to a uniform resolution of 640×640 pixels to ensure consistency across the dataset.

Annotations were carried out manually using COCO Annotator\cite{brooks:coco:2019}. Each image was annotated with instance segmentation masks that accurately delineate pill borders, even in overlap or partial occlusion instances. Multiple annotators worked on the dataset following strict inter-annotator guidelines designed to ensure consistency. Furthermore, each annotated image underwent manual examination by both the initial annotator and a secondary reviewer to detect and rectify any errors, guaranteeing a dependable, high-quality dataset.

\subsection*{MEDISEG (3-Pills)}

\begin{figure}[ht]
\centering
\includegraphics[width=\linewidth]{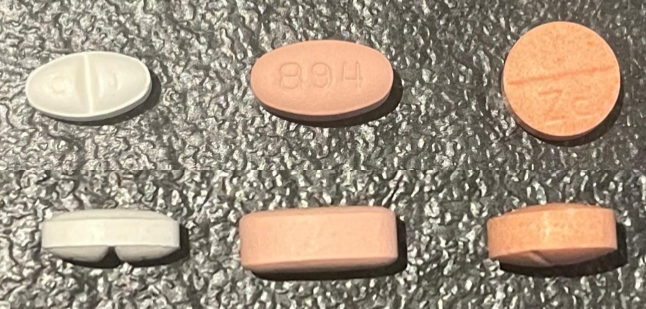}
\caption{Top-down and side profile views of Pill A (left), Pill B (centre), and Pill C (right).}
\label{fig:pills}
\end{figure}

\begin{figure}[ht]
\centering
\includegraphics[width=\linewidth]{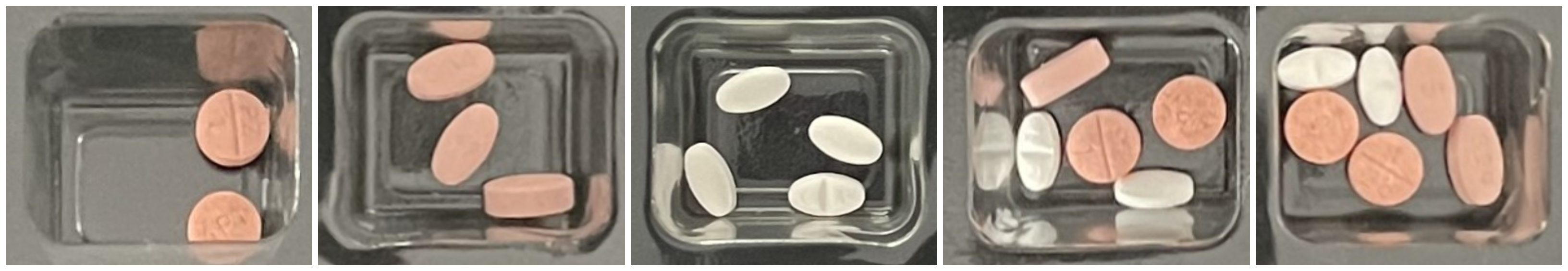}
\caption{Examples of image variability in the MEDISEG dataset, ranging from single-pill frames to compositions containing up to six pills. }
\label{fig:variability}
\end{figure}

\begin{figure}[ht]
\centering
\includegraphics[width=\linewidth]{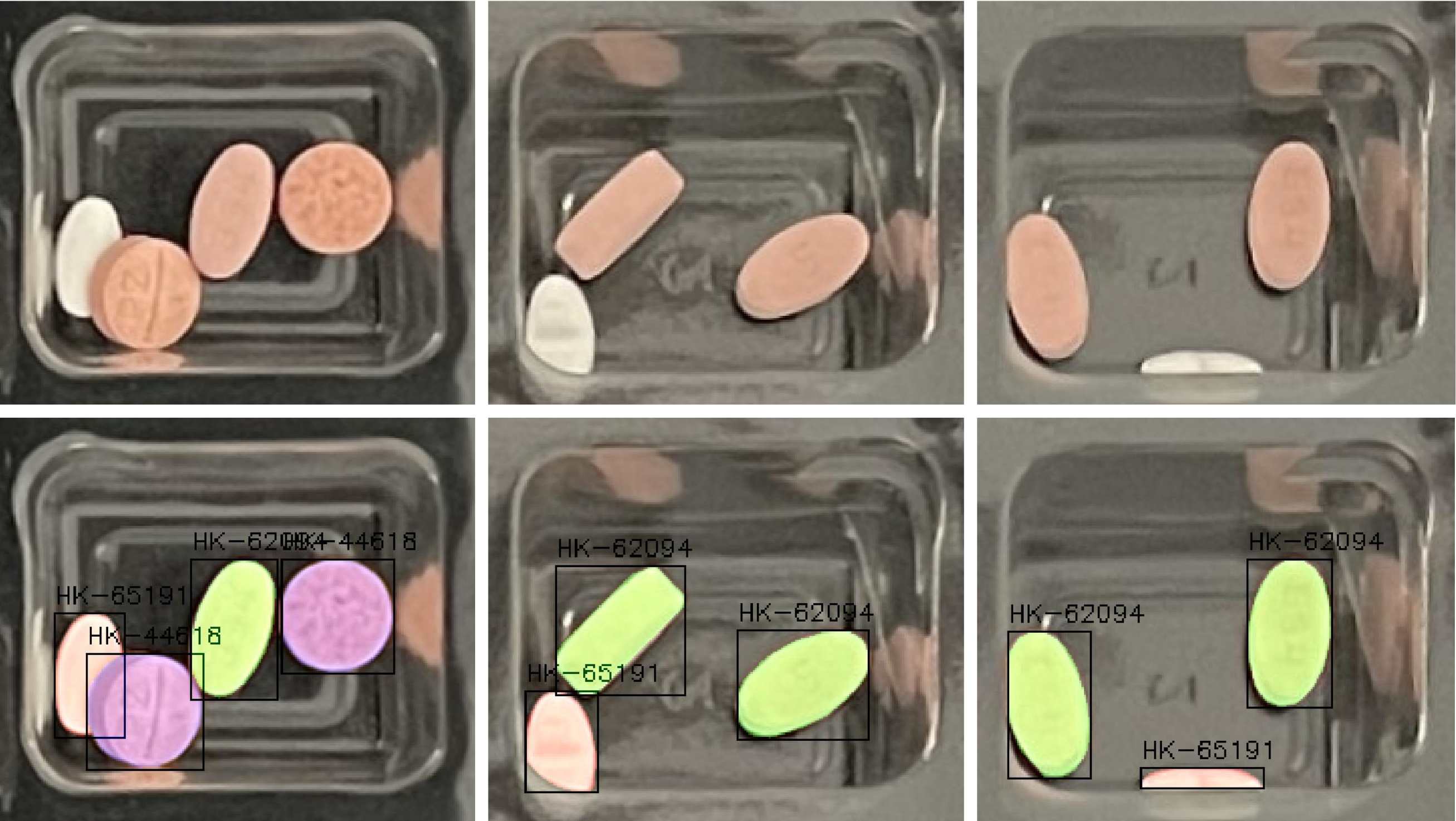}
\caption{Visualisation examples for the MEDISEG (3-Pills) dataset with instance masks, bounding boxes, and labels.}
\label{fig:visualise3}
\end{figure}

In its most basic form, the MEDISEG dataset contains three pill types deliberately chosen to challenge recognition algorithms. Figure \ref{fig:pills} shows the top-down view and side profile of the three pills selected—Pill A and Pill B share a comparable shape, which encourages the model to focus on colour features and fine-grained shape details, while Pill C exhibits a peach-like hue that closely resembles Pill B's colour, further testing the system's capacity to parse subtle chromatic differences. Moreover, as illustrated in Figure \ref{fig:variability}, the images span from single-pill frames to those depicting up to six pills, thereby incorporating diverse orientations, partial occlusions, and reflections. Finally, Figure \ref{fig:visualise3} presents visualisations with instance segmentation masks, bounding boxes, and labels superimposed on the images, clearly illustrating the diverse conditions a recognition model will likely face in practical scenarios.

\subsection*{MEDISEG (32-Pills)}
A more comprehensive version of the dataset expands to 32 different pill classes, spanning a broader spectrum of shapes, colours, and sizes. Crucially, the expansion also emphasises fine-grained, visually confusable cases, such as multiple small white tablets with near-identical geometry and finish, so that models must discriminate subtle inter-class differences rather than rely on coarse cues. This deliberate mix, together with scenes ranging from single pills to as many as 13 pills per frame and including occlusions and dosette-box edges, better mirrors home and clinical environments and provides a challenging benchmark for detection and segmentation.

\begin{figure}[ht]
\centering
\includegraphics[width=\linewidth]{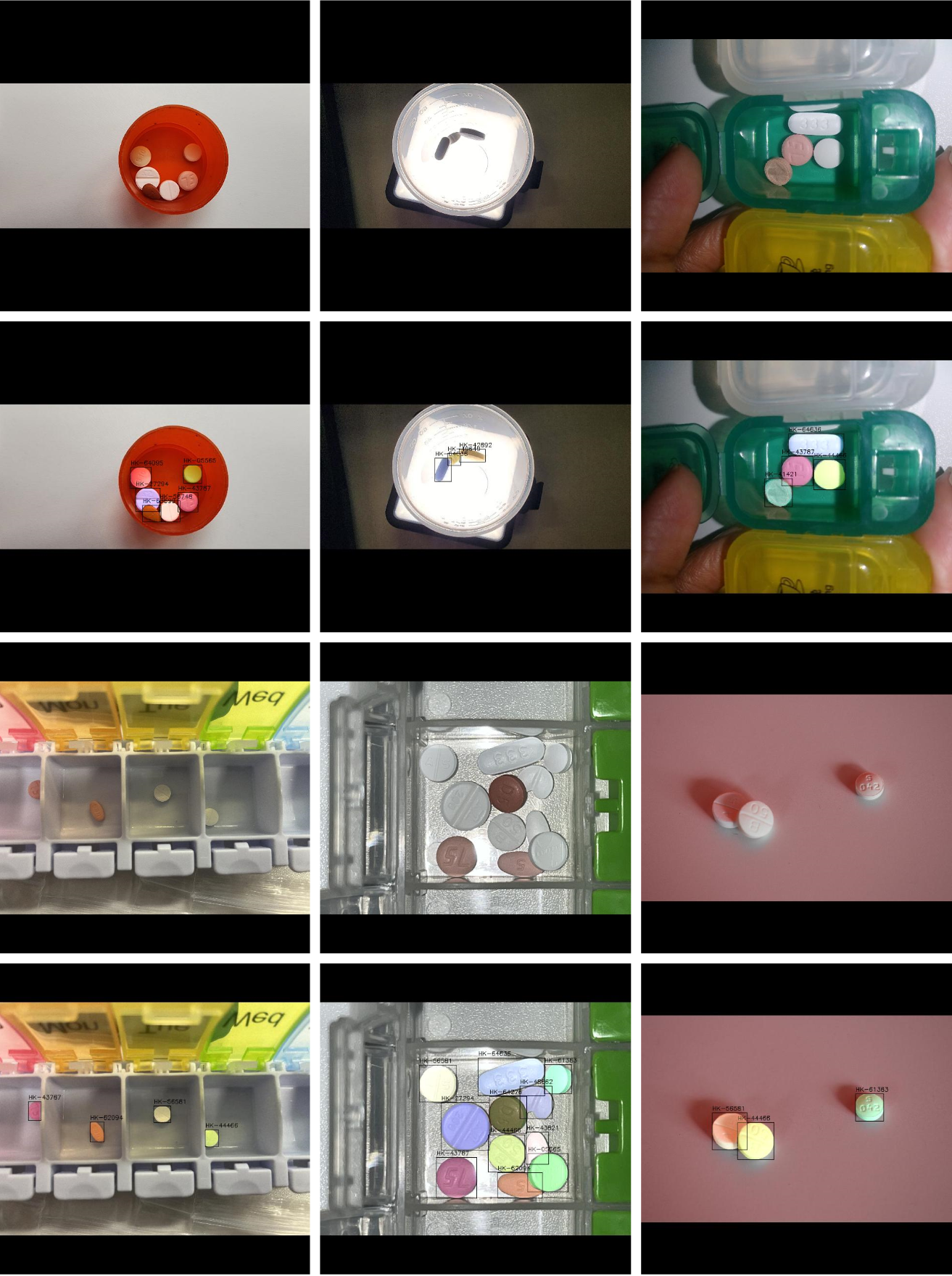}
\caption{Visualisation examples for the MEDISEG (32-Pills) dataset with instance masks, bounding boxes, and labels.}
\label{fig:visualise32}
\end{figure}

\begin{figure}[ht]
\centering
\includegraphics[width=\linewidth]{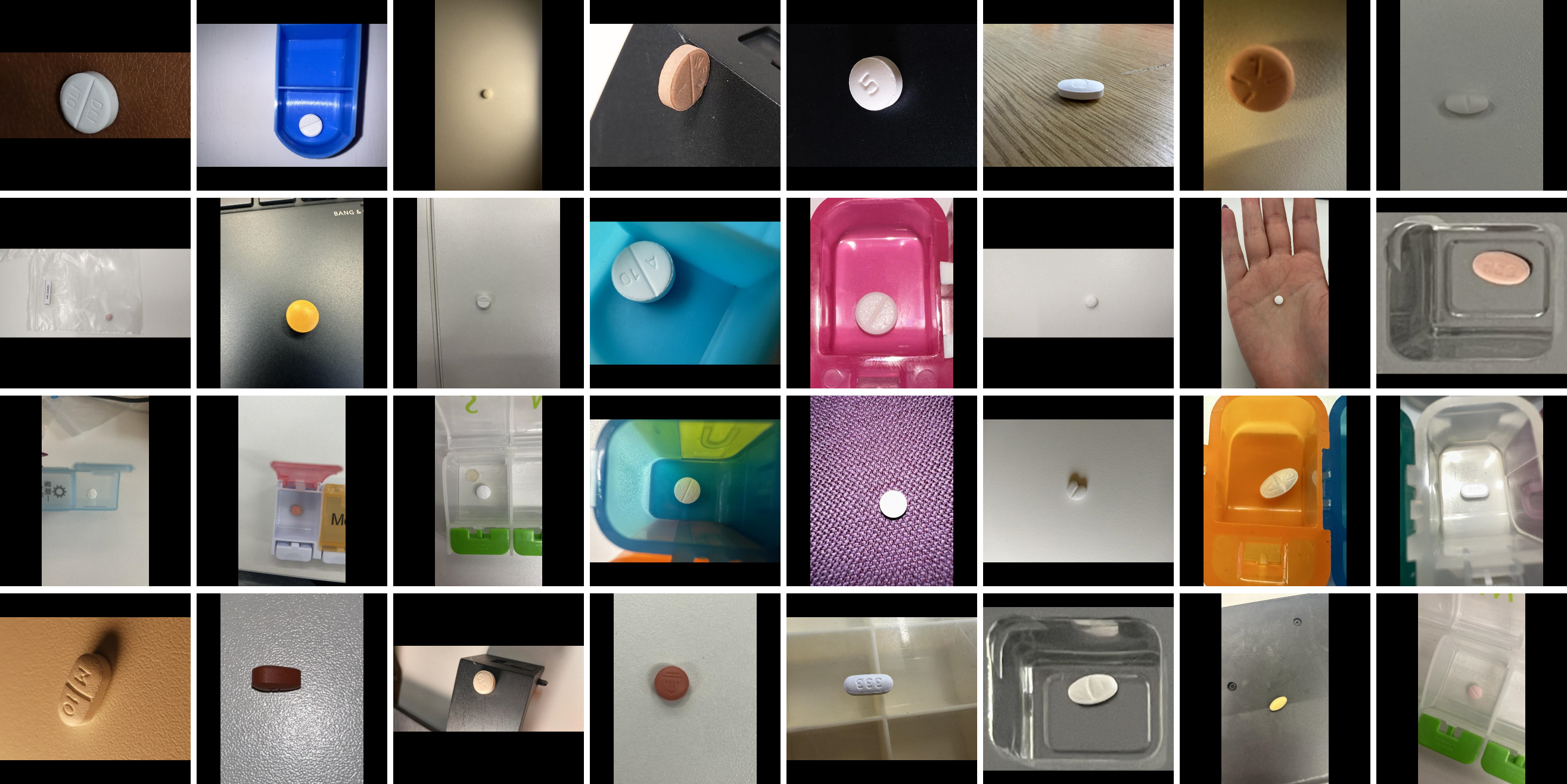}
\caption{Example images of the 32 unique pill types included in the MEDISEG (32-Pills) dataset.}
\label{fig:32pills_showcase}
\end{figure}

Each medication was photographed under varied lighting (intensity and angle), backgrounds, and with both single- and multi-pill compositions. We maintained per-class checklists to ensure coverage of (i) top-down and side views, (ii) occlusions/overlaps, (iii) presence of dosette walls/box edges, and (iv) specular highlights. These multi-pill images, coupled with different lighting conditions and heterogeneous background settings, place additional demands on models seeking to perform accurate pill recognition and segmentation. Figure \ref{fig:visualise32} provides examples from the MEDISEG (32-Pills) configuration, clearly showcasing the instance segmentation masks, bounding boxes, and class names. Figure~\ref{fig:32pills_showcase} provides a representative sample of the 32 unique pill types included in the dataset, showcasing the diversity in visual characteristics that models must learn to differentiate.

\subsection*{Spatial Distribution and Image Composition}

\begin{figure}[ht]
\centering
\includegraphics[width=\linewidth]{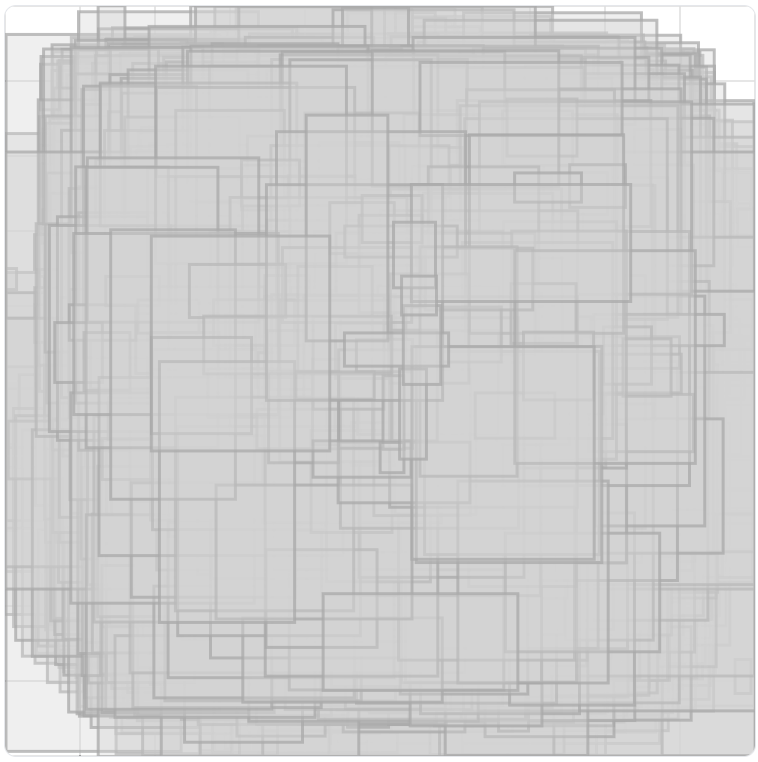}
\caption{Spatial distribution of bounding boxes across the MEDISEG (32-Pills) dataset. }
\label{fig:heatmap}
\end{figure}

\begin{figure}[ht]
\centering
\includegraphics[width=\linewidth]{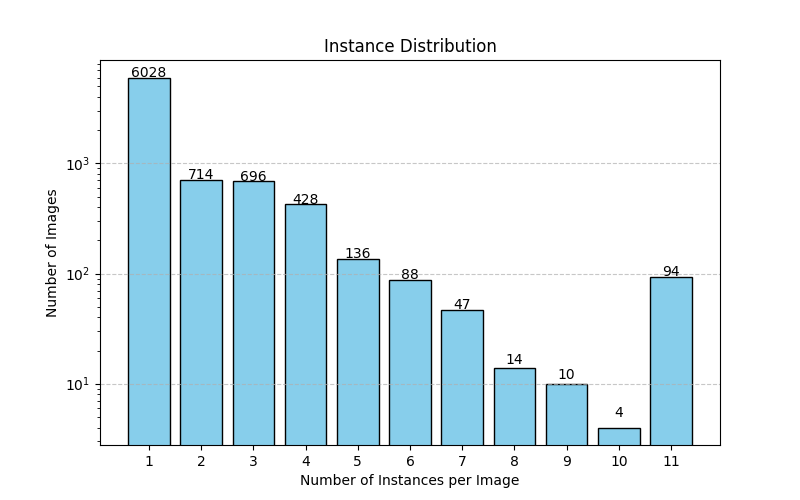}
\caption{Instance distribution of images in the MEDISEG dataset. }
\label{fig:instance_distribution}
\end{figure}

In order to evaluate and illustrate the variety within MEDISEG, two key analyses were conducted. First, an examination of spatial distribution, shown in Figure \ref{fig:heatmap}, highlights the locations of pill bounding boxes throughout the images. This visual overview demonstrates that pills are positioned across different regions of each frame rather than being consistently centered or clustered. Second, the histogram of annotation density in Figure \ref{fig:instance_distribution} illustrates the distribution of pill instances per image, ranging from single-pill images to those containing multiple pills. The x-axis represents the number of pill instances per image, spanning from 1 to 11, while the y-axis indicates the number of images corresponding to each instance count. A logarithmic scale is applied to the y-axis to better visualise the distribution. Through analysing these distributions, researchers can attain a more lucid comprehension of the dataset's intricacy, along with the potential obstacles that automated models might encounter when addressing overlapping or partially obscured pills and backdrop clutter.

Table \ref{tab:MEDISEG_dataset_comparison} compares the NIH Pillbox\cite{yaniv:nlm:2016}, CURE\cite{ling:few-shot:2020}, and MEDISEG datasets. While NIH Pillbox and CURE primarily consist of single-pill images with limited annotation detail, MEDISEG features multi-pill images with comprehensive instance segmentation labels, better reflecting real-world medication scenarios.

\subsection*{Few-shot protocol details}
To assess whether the visual complexity captured by MEDISEG contributes to improved recognition in realistic medication-handling scenarios, we conducted a limited-label evaluation using a few-shot object detection framework.

We adopted the FsDet (Few-shot Detection) framework, a widely used two-stage training paradigm for object detection under limited supervision. FsDet decomposes training into a base learning stage and a few-shot fine-tuning stage, allowing the effect of the base dataset’s visual characteristics to be isolated when only a small number of labelled examples are available for novel classes\cite{wang:FsDet:2020}.

The detector architecture is based on Faster R-CNN with a ResNet backbone and a Feature Pyramid Network (FPN)\cite{ren:RCNN:2017}. During base training, the model learns general visual representations, region proposals, and class-agnostic localisation using a set of base pill classes. Importantly, these base classes are disjoint from the novel classes used in the few-shot stage.

In the few-shot fine-tuning stage, the detector is adapted to previously unseen pill classes using a very limited number of annotated instances (1-, 5-, or 10-shot). To stabilise training under this extreme data constraint, the backbone feature extractor and most of the region proposal network are kept fixed. Only the region-of-interest heads responsible for classification and bounding box regression are fine-tuned, with the classification layer re-initialised to accommodate the novel classes. This training strategy is standard in FsDet-based evaluations and reduces overfitting when supervision is scarce.

Two base training configurations were compared in this study:
\begin{itemize}
    \item A model trained on the MEDISEG dataset, which includes real multi-pill scenes with frequent overlap, partial occlusion, and clutter; and
    \item A model trained on the CURE dataset, consisting primarily of isolated pill images captured under controlled conditions.
\end{itemize}

All other architectural components, optimisation settings, and fine-tuning schedules were kept identical to ensure a fair comparison.

Evaluation was conducted on a held-out set of multi-pill images annotated at the instance level. In addition to a standard test split, a challenging subset was constructed that contains only images with substantial pill overlap or occlusion, reflecting conditions commonly encountered in dosette boxes and medication trays.

\section*{Data Records}

The dataset is available at City St George's, University of London Figshare repository \cite{chu:mediseg:2025}. This section details the directory layout, file formats (COCO), and metadata fields.

The MEDISEG dataset is publicly available at \url{https://doi.org/10.25383/city.28574786.v1}, distributed under the Creative Commons Attribution 4.0 International (CC BY 4.0) license. The dataset's top-level directory is organised as follows:
\dirtree{%\
.1 MEDISEG/.
.2 LICENSE.
.2 metadata.csv.
.2 3pills/.
.3 annotations.json.
.3 images/.
.4 image1.jpg.
.4 image2.jpg.
.2 32pills/.
.3 annotations.json.
.3 images/.
.4 image1.jpg.
.4 image2.jpg.
}

\texttt{LICENSE} contains the complete text of the CC BY 4.0 license under which the dataset is distributed.
The spreadsheet named \texttt{metadata.csv} provides supplementary information about the medications. Each row represents a single drug entry using the following columns:

\begin{itemize}
\item id: The official Hong Kong drug registration number (e.g., HK-05565).
\item name: The brand name and dosage strength of the medication (e.g., “LASIX TAB 40MG”).
\item certificate\_holder: The company or entity holding the product certificate.
\item certificate\_holder\_address: Official address of the certificate holder, often relevant for regulatory or traceability purposes.
\item ingredients/0, ingredients/1: One or two active ingredients listed by generic name.
\item sale\_requirement: Classification code indicating whether the medication is prescription-only or subject to other sales restrictions (0 for Over-the-Counter Medicines, 1 for Pharmacy Only Medicines, 2 for Prescription Only Medicines).
\item registration\_date: The date on which the drug received its local registration.
\item url: A direct link to the Hong Kong Drug Office's product detail page, providing further regulatory and labelling information.
\end{itemize}

3pills and 32pills directories contain the MEDISEG (3-Pills) and MEDISEG (32-Pills) subsets, respectively. Data are further organised as follows:

\begin{itemize}
\item annotations.json: A COCO-format file that describes every labelled pill instance: its bounding box coordinates, segmentation polygons, and class labels. 
\item images: A folder containing all JPG files. 
\end{itemize}

\section*{Technical Validation}

To evaluate the quality and usability of the MEDISEG dataset, we conducted experiments using two state-of-the-art object detection models, YOLOv8\cite{ultralytics:yolov8:2023} and YOLOv9\cite{wang:yolov9:2024}, on both the MEDISEG (3-Pills) and MEDISEG (32-Pills) datasets. Model training was performed using a hold-out method, with 70\% of the data used for training, 20\% for validation, and 10\% reserved for testing. The reported results are relative to the validation set. 

Model performance was evaluated using standard metrics, including precision, recall, F1-score, mAP@50, and mAP@50-95. Precision measures the proportion of correctly identified pill instances among all predicted instances and is defined as:
\[
\text{Precision} = \frac{\sum_{i=1}^{n} TP_i}{\sum_{i=1}^{n} (TP_i + FP_i)}
\]
Where \textit{TP} represents objects correctly classified as positive and \textit{TP}+\textit{FP} represents everything classified as positive. In the context of pill recognition, precision measures how well the model avoids incorrectly identifying pills as a class they do not belong to. 

Similarly, recall quantifies the model's ability to detect all actual pill instances and is defined as:
\[
\text{Recall} = \frac{\sum_{i=1}^{n} TP_i}{\sum_{i=1}^{n} (TP_i + FN_i)}
\]
Where \textit{TP} represents objects correctly classified as positive, and \textit{TP}+\textit{FN} represents all actual positive objects. In the context of pill recognition, recall measures how well the model identifies all instances of a particular pill class, ensuring that no pills are missed during detection. 

The F1-Score provides the harmonic mean of precision and recall, offering a balanced metric that considers both FP and FN. It is defined as:
\[
\text{F1-Score} = 2\cdot\frac{\text{Precision}\cdot\text{Recall}}{\text{Precision}+\text{Recall}}
\]

To evaluate overall detection performance, we compute Average Precision (AP), which represents the area under the precision-recall curve. It is computed as:
\[
AP = \frac{1}{11} \sum_{r \in \{0, 0.1, 0.2, \dots, 1\}} P_{\text{interp}}(r)
\]
Where \textit{r} represents the recall levels at 0, 0.1, 0.2, ..., 1. For each recall level \textit{r}, the maximum precision observed at or above \textit{r} is considered:
\[
P_{\text{interp}}(r) = \max_{\tilde{r} \geq r} \text{Precision}(\tilde{r})
\]

Mean Average Precision (mAP) extends AP across multiple classes and provides a single metric to assess model performance across all pill categories: 
\[
mAP = \frac{1}{N}\sum_{i=1}^{N}AP_i
\]
where \textit{N} represents the total number of classes. In the context of pill recognition, mAP evaluates the model's effectiveness in detecting and classifying all pill classes. mAP@50 captures the model's general detection ability, while mAP@50-95 assesses its robustness in handling challenging scenarios, such as overlapping or occluded pills. A high mAP value indicates the model's reliability in real-world healthcare applications, where precise and consistent pill recognition is critical.

We present comprehensive results for each dataset configuration, compare the performance of YOLOv8 and YOLOv9, and analyse the implications of these findings for practical pill detection tasks.

A high-level overview of the YOLOv8 and YOLOv9 architectures highlights that both models follow the standard YOLO detection pipeline. Each architecture comprises a backbone network for feature extraction, a central module to fuse multi-scale features, and specialised heads that output bounding box coordinates and class probabilities. While YOLOv9 builds upon YOLOv8 with incremental improvements for small-object detection and refined feature fusion, both models serve as versatile, high-performance baselines for object detection tasks.

\subsection*{Hyperparameter Tuning}
\begin{figure}[ht]
\centering
\includegraphics[width=\linewidth]{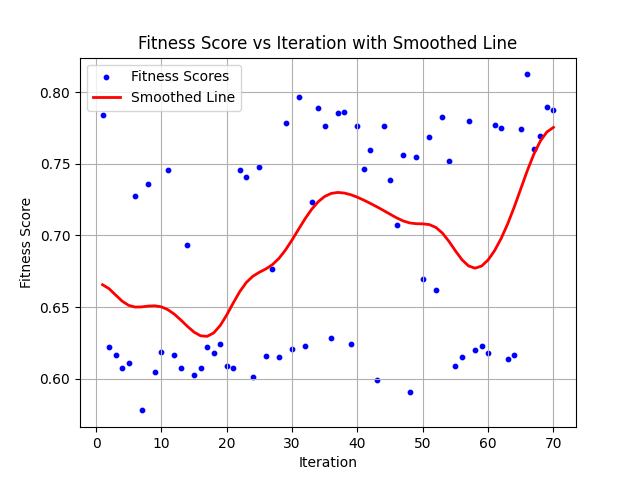}
\caption{Fitness score progression over 70 iterations of hyperparameter tuning using GA}
\label{fig:tune_results}
\end{figure}

A hyperparameter tuning approach utilising a genetic algorithm (GA) was performed over 70 iterations, concentrating on optimising key parameters such as learning rate, momentum, and weight decay to improve the performance of the YOLO model. Figure \ref{fig:tune_results} illustrates the progression of the fitness score across iterations. The results demonstrate a steady improvement in the fitness score as the GA converged towards more optimal hyperparameter configurations. Notably, the initial iterations exhibited a wide range of fitness scores, reflecting the exploratory nature of GA. However, as the iterations progressed, the fitness scores stabilised and clustered around higher values.

The fitness score is a metric used to evaluate the performance of each hyperparameter configuration during the tuning process. It is calculated as the mean of two metrics: top-1 accuracy and top-5 accuracy. The top-1 accuracy measures the percentage of predictions where the model's highest confidence class matches the true label, offering a strict evaluation of prediction correctness. In contrast, top-5 accuracy accounts for the percentage of predictions where the true label is among the model's top five predicted classes, providing a broader evaluation, particularly useful for multi-class classification tasks with highly similar classes. The fitness score is computed as follows:
\[
\text{Fitness Score} = \frac{\text{Top-1 Accuracy} + \text{Top-5 Accuracy}}{2}
\]

By averaging these two metrics, the fitness score balances the model's ability to make confident predictions while capturing its capacity to generalise across multiple plausible outcomes. This ensures a comprehensive assessment of model performance. The highest fitness score achieved during the tuning process was 0.81253 in iteration 66. This was achieved with the following set of hyperparameters:
\begin{itemize}
  \item Learning rate (lr0): 0.01009
  \item Learning rate final (lrf): 0.01
  \item Momentum: 0.94023
  \item Weight decay: 0.00048
  \item Warmup epochs: 2.9974
  \item Warmup momentum: 0.66015
  \item Box loss weight: 7.26074
\end{itemize}

\subsection*{Validation on MEDISEG (3-Pills)}

The MEDISEG (3-Pills) subset serves as a foundational testbed for evaluating model performance under moderately complex conditions. It comprises three pill types, two of which share a similar shape (Pill A and Pill B), while Pill B and Pill C closely resemble each other in colour. This setup challenges the models' ability to distinguish subtle differences in both shape and hue.

\begin{figure}[ht]
\centering
\includegraphics[width=\linewidth]{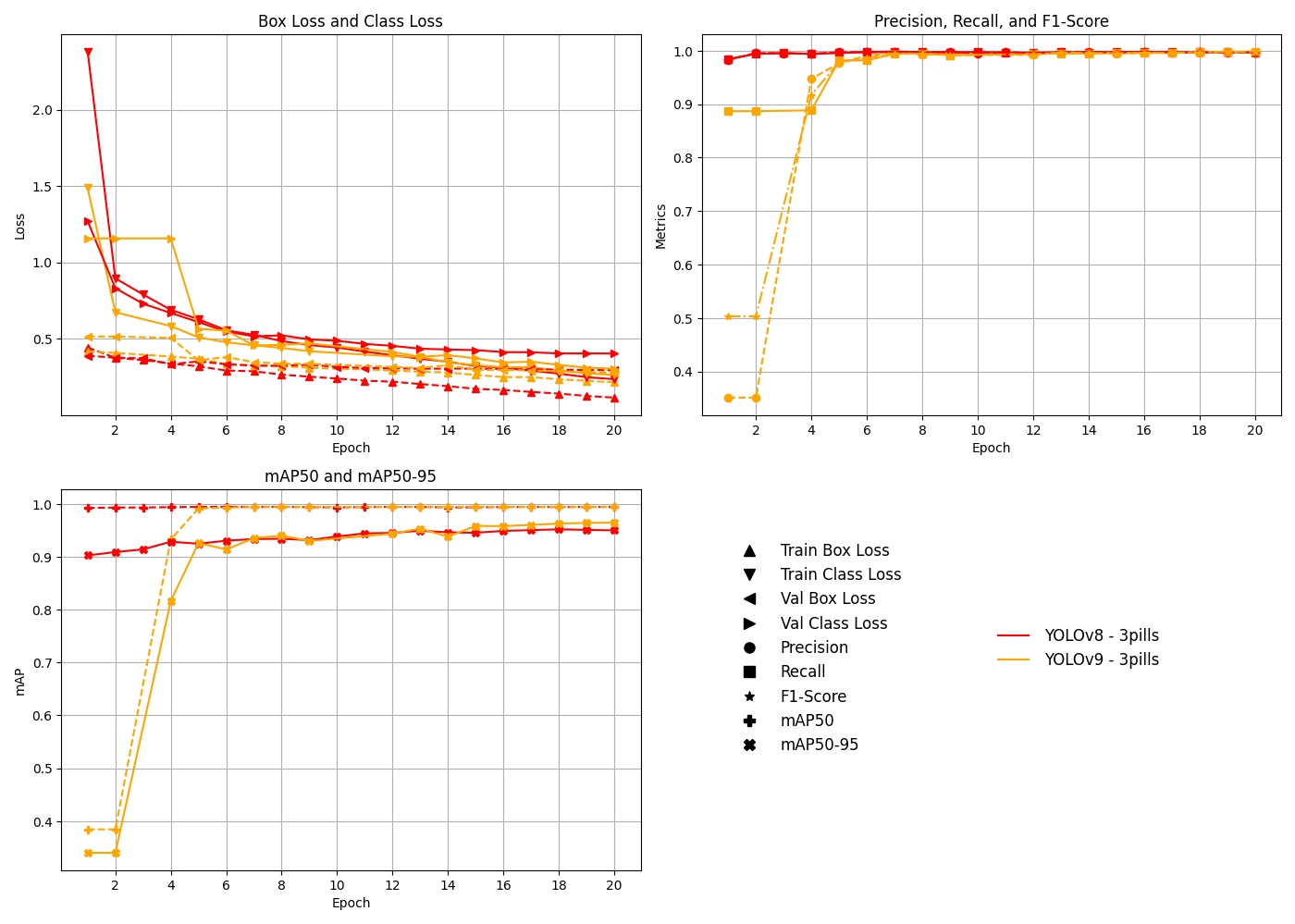}
\caption{Comparison of YOLOv8 and YOLOv9 training metrics on the MEDISEG (3-Pills) dataset.}
\label{fig:results_3}
\end{figure}

Figure \ref{fig:results_3} presents the training and validation curves for YOLOv8 and YOLOv9 on the 3-Pills dataset. Both models demonstrate robust training behaviour, characterised by a gradual and consistent decrease in box loss and class loss throughout the epochs. YOLOv8 exhibits fast convergence, attaining low loss values early in the training process, whereas YOLOv9 displays a comparable pattern but enhances more gradually. Although it exhibits a slower starting phase, YOLOv9 outperforms YOLOv8 in mAP@50-95, underscoring its enhanced localisation capabilities under more stringent Intersection over Union (IoU) thresholds.

Across the training process, precision and recall curves indicate that both models maintain high detection reliability. YOLOv8 achieves strong initial performance, with precision and recall stabilising near 1.0 early on. In contrast, YOLOv9 exhibits a sharp increase in recall and precision after the first few epochs, suggesting an efficient learning phase. By the end of the training, both models converged to high values of mAP, confirming their robustness in handling the 3-Pills dataset.

For a comprehensive comparison of final model performance, including precision, recall, and F1-score metrics, refer to Table \ref{tab:YOLOv8_3} and Table \ref{tab:YOLOv9_3}.

\begin{figure}[ht]
\centering
\includegraphics[width=\linewidth]{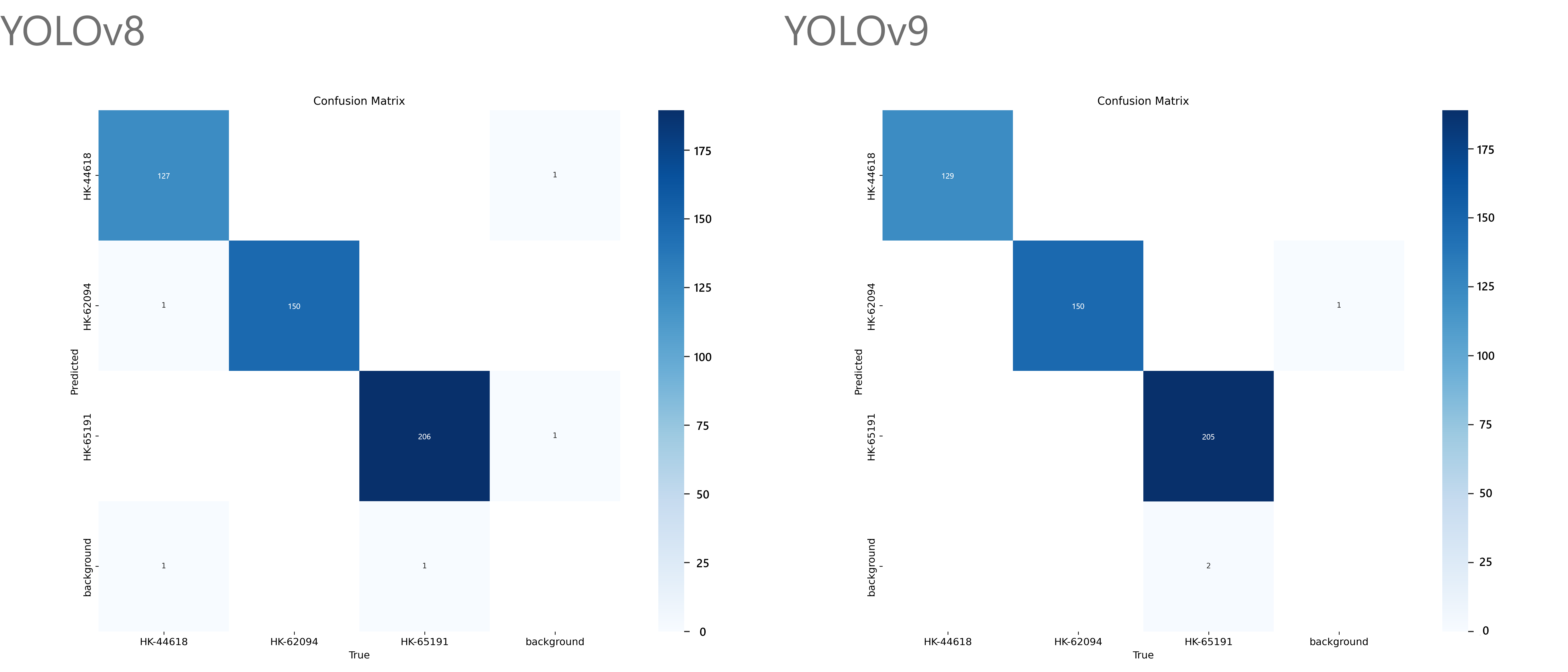}
\caption{Confusion matrices for YOLOv8 (left) and YOLOv9 (right) trained on the MEDISEG (3-Pills) dataset, evaluated on the test set.}
\label{fig:confusion_3}
\end{figure}

To assess generalisation, we evaluated both models on the test set, which was not used at any point during training or validation, ensuring truly unseen data. The confusion matrices in Figure \ref{fig:confusion_3} illustrate the classification performance across the three pill classes. Both models maintain high true positive rates, while YOLOv9 exhibits a marginally lower false positive rate compared to YOLOv8.

\begin{figure}[ht]
\centering
\includegraphics[width=\linewidth]{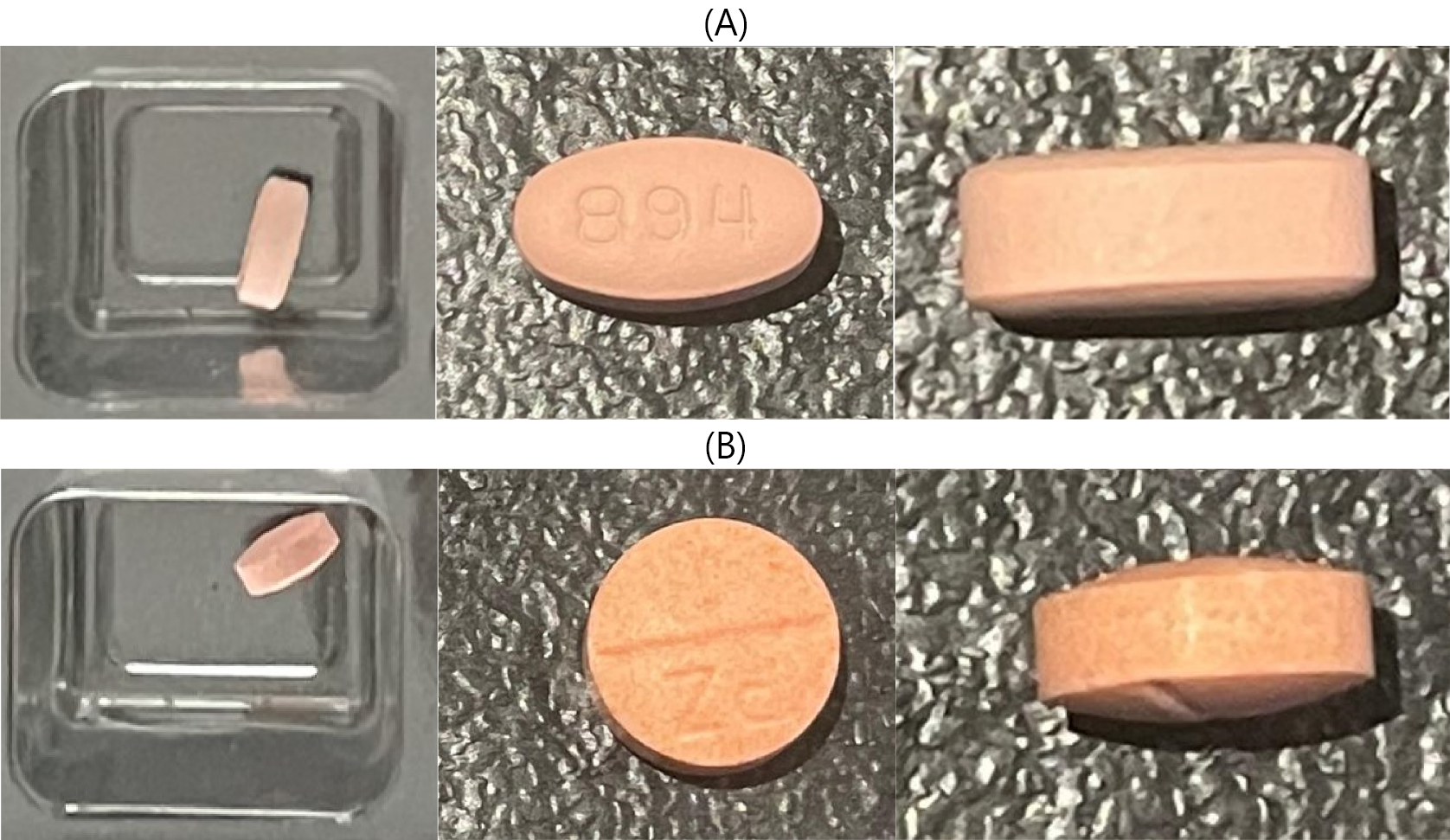}
\caption{Common misclassifications: (A) Pill B and (B) Pill C. Left: Sample image; Middle: Top-down view; Right: Side profile.}
\label{fig:misclassification}
\end{figure}

\begin{figure}[ht]
\centering
\includegraphics[width=\linewidth]{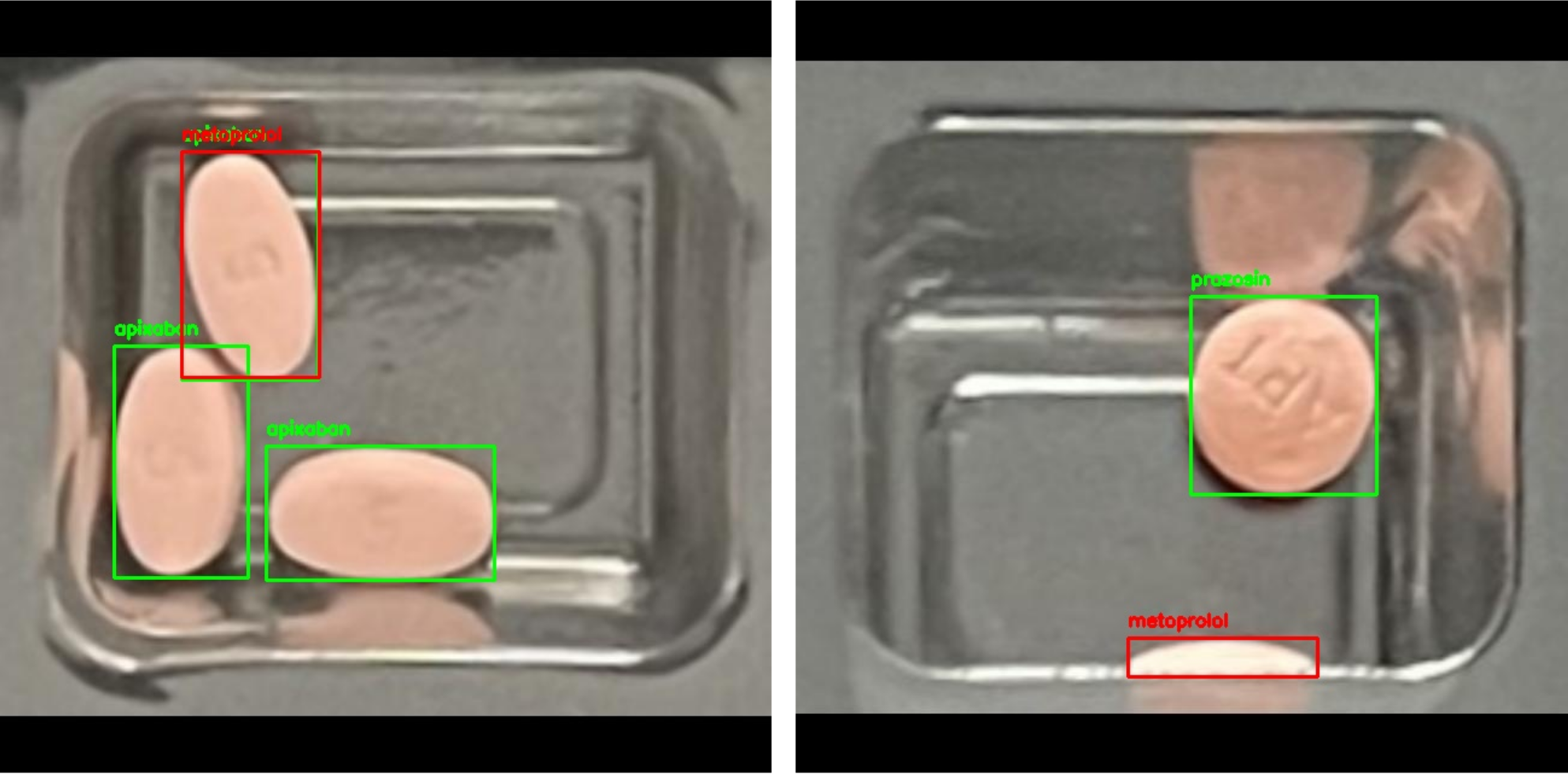}
\caption{Misclassification examples}
\label{fig:misclassification_example}
\end{figure}

Misclassifications predominantly stemmed from nuanced colour variations between Pill B and Pill C. As depicted in Figure \ref{fig:misclassification}, while the top-down view highlights distinct differences that make the pills easily distinguishable, their side profiles appear strikingly similar, often leading to errors in classification. Figure \ref{fig:misclassification_example} presents two examples of these misclassifications. In these images, green bounding boxes indicate correct classifications, while red bounding boxes highlight errors. Both YOLOv8 and YOLOv9 struggled in these edge cases, though YOLOv8 exhibited slightly higher misclassification rates overall.

\subsection*{Validation on MEDISEG (32-Pills)}

The more comprehensive MEDISEG (32-Pills) subset expands the range of pill categories and real-world scenarios. It includes 32 different pill classes with varied shapes, colours, and sizes, often presented in images containing multiple pills under diverse lighting and background conditions. This dataset introduces a greater level of complexity, making it a more challenging benchmark for model evaluation.

\begin{figure}[ht]
\centering
\includegraphics[width=\linewidth]{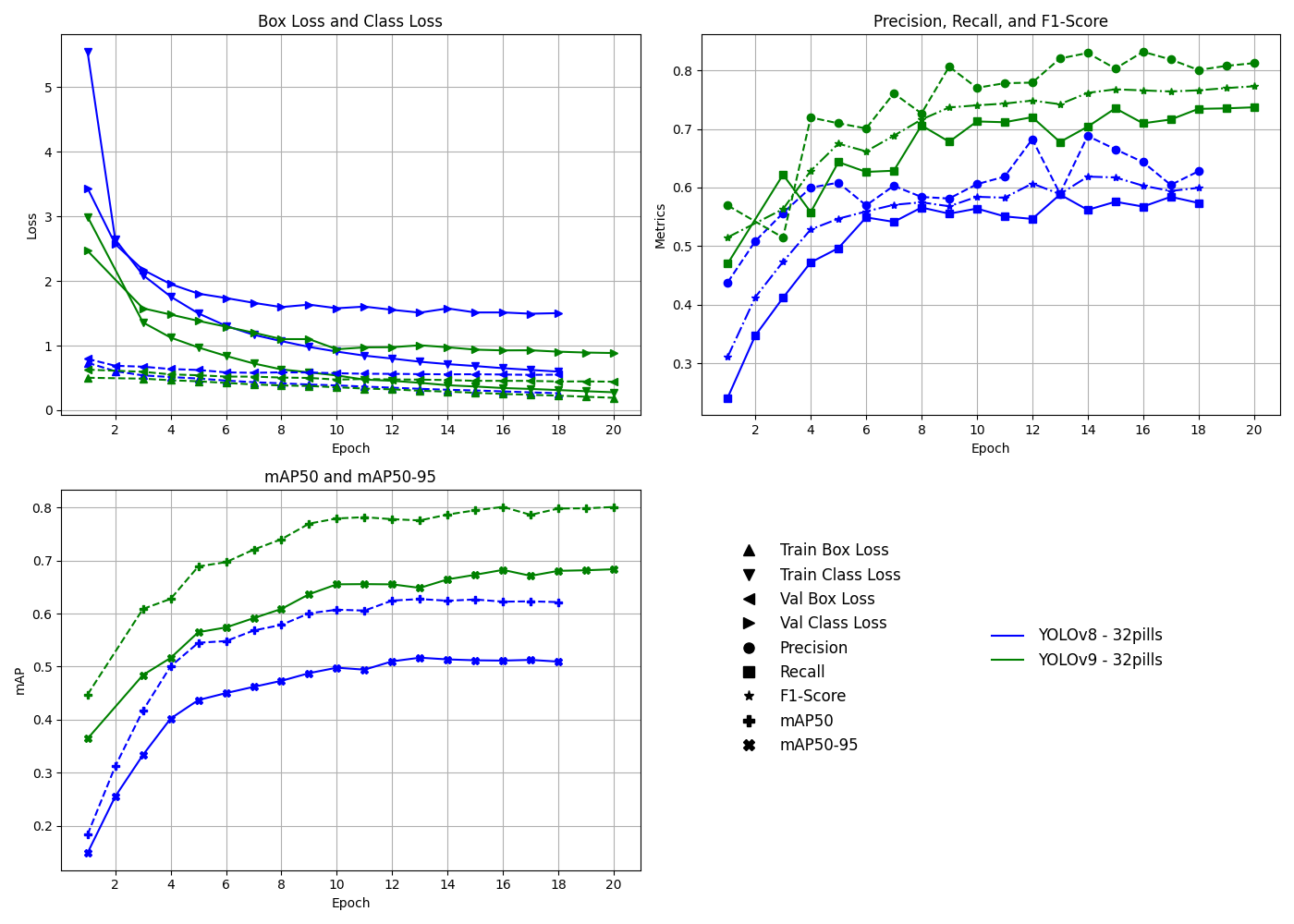}
\caption{Comparison of YOLOv8 and YOLOv9 training metrics on the MEDISEG (32-Pills) dataset.}
\label{fig:results_32}
\end{figure}

Figure \ref{fig:results_32} shows the training progression for YOLOv8 and YOLOv9 on the 32-Pills dataset. While both models demonstrate stable learning behaviour, YOLOv8 converges more rapidly in the early epochs, mirroring its behaviour on the 3-Pills subset. However, as training progresses, YOLOv9 steadily surpasses YOLOv8 in mAP@50-95, highlighting its stronger capability for precise localisation across a wider variety of pill classes. This trend suggests that while YOLOv8 benefits from fast deployment, YOLOv9 may be the better choice for applications requiring fine-grained pill recognition.

Precision and recall curves further reinforce this pattern—YOLOv8 attains high precision earlier, while YOLOv9 exhibits a more gradual yet consistent improvement over epochs. The final mAP values confirm strong performance for both models, with YOLOv9 achieving a higher mAP@50-95, demonstrating its superior ability to handle occlusions, overlapping pills, and more diverse object classes.

For detailed performance metrics, including precision, recall, and F1-score comparisons, refer to Table \ref{tab:YOLOv8_32} and Table \ref{tab:YOLOv9_32}.

\begin{figure}[ht]
\centering
\includegraphics[width=\linewidth]{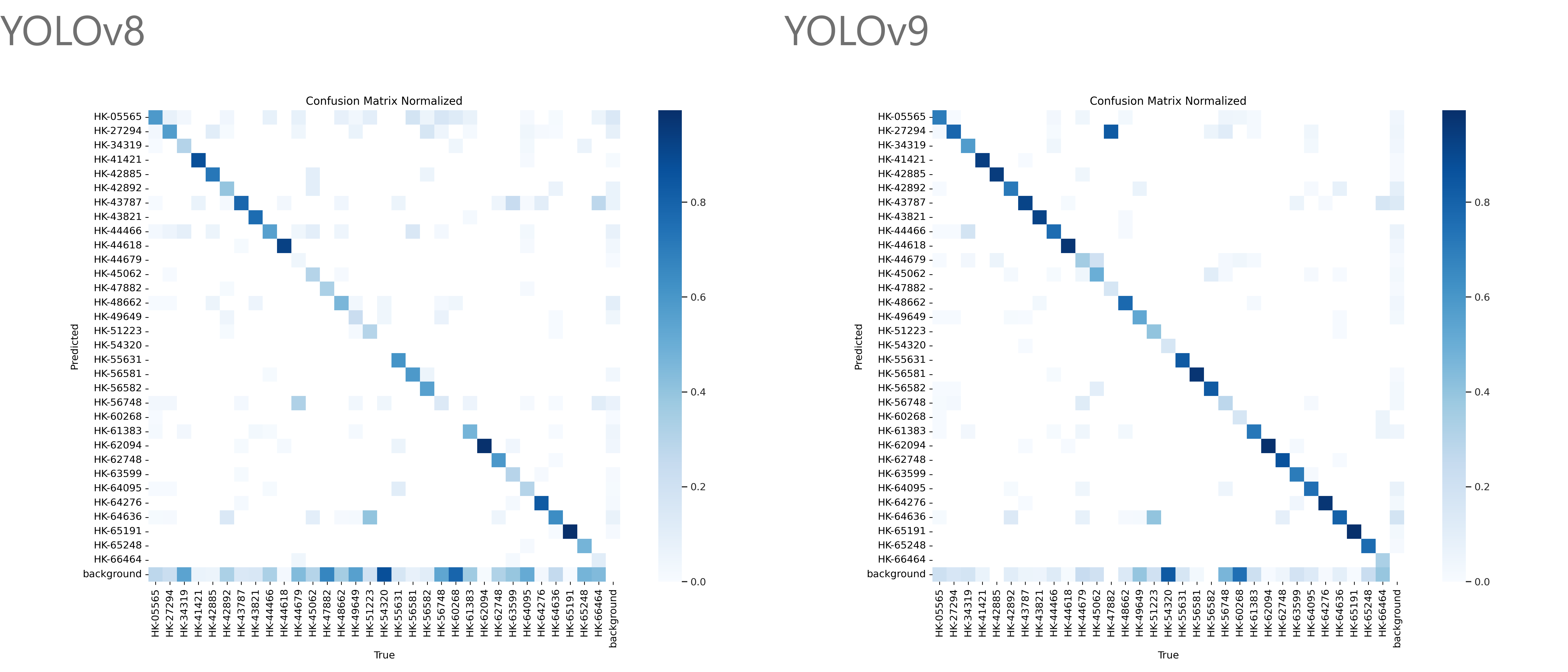}
\caption{Confusion matrices for YOLOv8 (left) and YOLOv9 (right) trained on the MEDISEG (32-Pills) dataset, evaluated on the test set.}
\label{fig:confusion_32}
\end{figure}

To assess real-world generalisation, both models were evaluated on a test set that was never used during training or validation. The confusion matrices in Figure \ref{fig:confusion_32} highlight that YOLOv9 reduces false positives and false negatives more effectively than YOLOv8, particularly for visually similar or partially occluded pills.

\begin{figure}[ht]
\centering
\includegraphics[width=\linewidth]{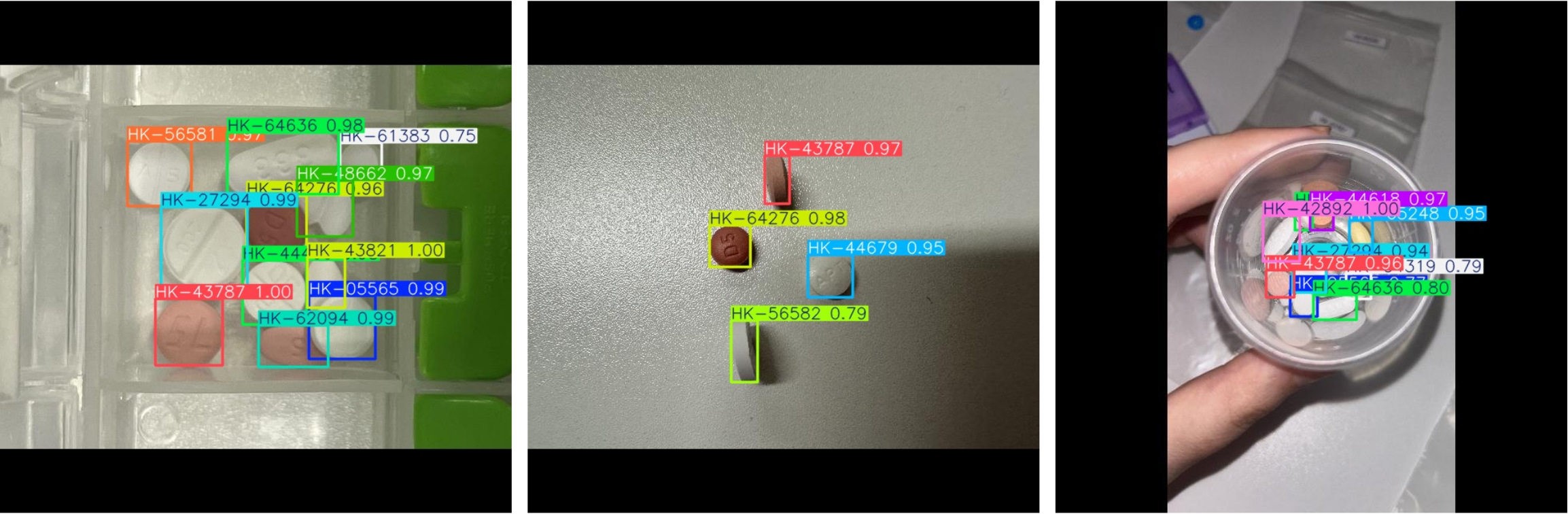}
\caption{Examples of model prediction on 32-Pills test images. }
\label{fig:32pill_pred}
\end{figure}

Additionally, Figure \ref{fig:32pill_pred} presents qualitative examples of model predictions on test images, illustrating bounding boxes, class labels, and confidence scores. These visualisations further confirm that both models perform well on unseen data.

\subsection*{Few-shot generalisation on unseen pill types}
In all few-shot configurations, models trained on MEDISEG base classes consistently outperformed those trained on the comparison dataset when evaluated on complex multi-pill scenes.

On the standard evaluation set, both training configurations achieved high foreground classification accuracy, even in the 1-shot setting. This result indicates that basic pill appearance can be learnt from a small number of examples. However, this evaluation did not strongly differentiate between datasets because the task is dominated by clear visual cues.

Table~\ref{tab:overlap} summarises few-shot detection performance on the overlap-only test set. Performance differences became pronounced when evaluation was restricted to images containing overlapping or partially occluded pills. In the 1-shot setting, the model base-trained on MEDISEG achieved a foreground classification accuracy of 0.406, compared to 0.131 for the model trained on the CURE dataset, while the false negative rate was also substantially reduced, reflecting improved instance discovery in visually ambiguous regions.

This pattern persisted as the number of labelled examples increased. In the 5-shot configuration, the MEDISEG-trained model achieved a foreground classification accuracy of 0.625, compared to 0.372, while in the 10-shot setting, accuracy increased to 0.740 versus 0.558. Across all few-shot configurations, base training on MEDISEG was associated with lower region proposal and total detection losses, indicating more stable localisation and classification behaviour in visually complex scenes.

\begin{figure}[ht]
\centering
\includegraphics[width=\linewidth]{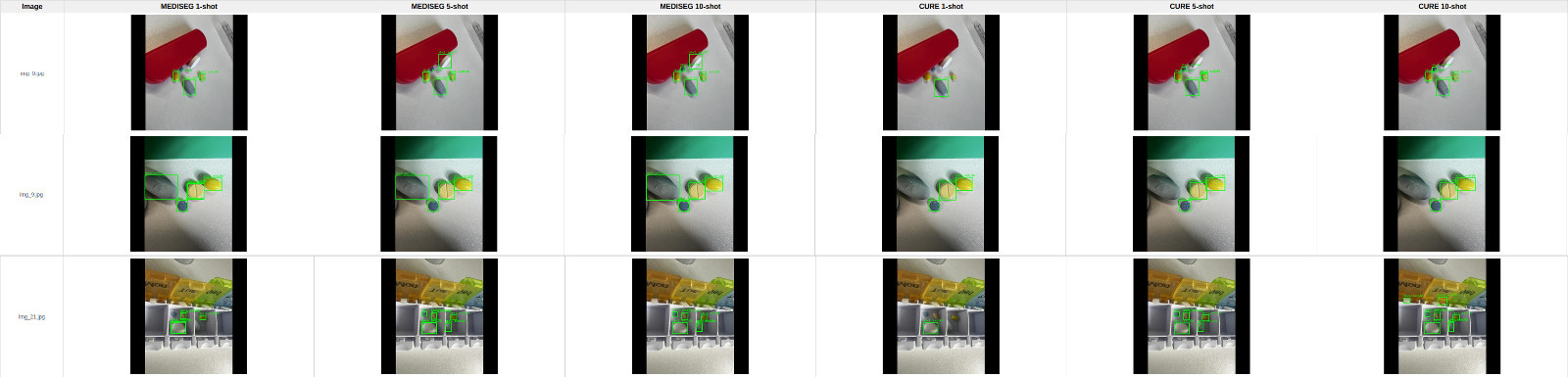}
\caption{Detection results under 1-shot, 5-shot, and 10-shot fine-tuning for models initialised from MEDISEG and CURE. }
\label{fig:fsl_vis}
\end{figure}

Figure \ref{fig:fsl_vis} presents a qualitative comparison of detection outputs under 1-, 5-, and 10-shot fine-tuning. Models initialised with MEDISEG demonstrate more consistent separation of overlapping instances and fewer missed detections, particularly under minimal supervision. While both models improve with additional labelled examples, MEDISEG initialisation maintains clearer object boundaries and more reliable classification in visually complex arrangements.

These results indicate that exposure to realistic multi-object interactions during base training substantially improves recognition performance in challenging pill arrangements, particularly when only limited annotated data are available. The improvements are most pronounced in scenes with heavy overlap and occlusion, which are underrepresented in existing pill image datasets but are common in real-world medication handling.

Our experiments demonstrate that MEDISEG, in both its focused (3-Pills) and comprehensive (32-Pills) configurations, supports stable training of modern object detection architectures such as YOLOv8 and YOLOv9. High mAP scores across both subsets confirm that MEDISEG’s realistic imaging conditions and dense instance-level annotations enable effective supervised learning in multi-pill scenarios.

Beyond fully supervised evaluation, our few-shot experiments further establish the dataset’s utility. Using a controlled FsDet protocol with base/novel class separation, we show that base training on MEDISEG substantially improves adaptation to unseen pill classes. Performance gains are most pronounced on the overlap-only test subset, where models pre-trained on MEDISEG consistently achieve higher foreground classification accuracy and lower false negative rates across 1-, 5-, and 10-shot configurations compared to models trained on a more controlled comparison dataset.

These findings indicate that exposure to realistic multi-object interactions during base training leads to more transferable representations, particularly for localisation and discrimination in visually cluttered scenes. The performance gap under heavy occlusion—conditions that closely resemble dosette boxes and dispensing trays—suggests that MEDISEG captures structural visual complexities not well represented in prior pill datasets. This structural realism appears to meaningfully improve generalisation under limited supervision.

From an application perspective, these results are relevant to medication verification workflows in both clinical and home settings. Bedside administration and dosette preparation frequently involve multiple pills arranged in close proximity, often partially occluded or rotated. In such environments, the ability to adapt to newly introduced formulations with limited labelled data is critical. Our few-shot evaluation under occlusion simulates this constraint, and the observed reduction in missed detections suggests that MEDISEG-based pretraining may enhance reliability in realistic multi-pill handling scenarios.

While these experiments demonstrate dataset utility and representation transfer, they do not constitute clinical validation. Prospective evaluation in live healthcare or pharmacy environments would be required prior to deployment, particularly to assess performance across broader device variability, lighting conditions, and workflow integration constraints.

Finally, the dataset’s emphasis on visually similar pill classes and multi-instance compositions provides a challenging benchmark for emerging architectures targeting fine-grained object discrimination and robust localisation. Future work may extend MEDISEG to include additional pill types, diversified acquisition devices, and evaluation under semi-supervised, continual learning, or domain adaptation settings to further explore its applicability to evolving healthcare AI systems.

\section*{Usage Notes}

Researchers can load and use the MEDISEG dataset by taking advantage of widely used computer vision libraries and APIs. Both 3pills/annotations.json and 32pills/annotations.json adhere to the standard COCO format, ensuring direct interoperability with frameworks like Detectron2, PyTorch, and TensorFlow. In particular, users can employ the official COCO API (available on GitHub) to parse the annotation files, interact with bounding box and segmentation data, and quickly set up training or evaluation pipelines. For those seeking to visualise or modify the existing labels, open-source tools such as LabelMe, CVAT, or COCO Annotator support COCO-format annotations and provide straightforward interfaces for overlaying segmentation masks on corresponding images.

The supplementary \texttt{metadata.csv} file offers insight into each pill's regulatory background, linking drug images to their marketing authorisation details, ingredient information, and official registration records. Researchers interested in exploring associations between visual characteristics and specific ingredients or dosage forms can merge this metadata with annotation data to investigate potential correlations or to support downstream tasks like automated drug identification. 

Neither specialised hardware nor proprietary software is required for most dataset applications. Standard computing resources suffice for parsing, visualising, or performing basic analyses on smaller subsets, while GPU-accelerated systems become more relevant if training large-scale deep learning models. When redistributing or referencing this dataset, please cite it in accordance with the CC BY 4.0 license and refer to the Methods section for additional details on data collection, preprocessing steps, and annotation protocols.

\section*{Code Availability}

All code used in this study was written in Python 3 and is provided as Jupyter Notebooks on GitHub at \url{https://github.com/williamcwi/MEDISEG}. The notebooks rely on the Ultralytics library for YOLOv8 and YOLOv9 implementations alongside standard Python packages. Version details and a complete list of dependencies are outlined in the repository's requirements.txt. 

\section*{Acknowledgements}

We gratefully acknowledge the City St George's Master Student for their invaluable assistance during the image creation and annotation process. 

\section*{Author Contributions}

\textbf{Ling Li:} Conceptualization, Supervision, Methodology, Study Design, Project Administration, Validation, Investigation, Resources, Writing – Review \& Editing. \textbf{Giacomo Tarroni:} Supervision, Validation, Investigation, Methodology, Writing – Review \& Editing. \textbf{Shashi Hirani:} Writing – Review \& Editing. \textbf{Wai Ip Chu:} Conceptualization, Methodology, Data Curation, Formal Analysis, Writing - Original Draft, Visualization. 

\section*{Competing Interests}

The author(s) declare no competing interests.

\bibliography{bibliography}

\clearpage
\section*{Figures}

% \begin{figure}[ht]
% \centering
% \includegraphics[width=\linewidth]{img/3pill_pred.PNG}
% \caption{Examples of model prediction on 3-Pills test images. }
% \label{fig:3pill_pred}
% \end{figure}

\clearpage
\section*{Tables}

\begin{table}[ht]
    \centering
    \begin{tabular}{|l|l|l|l|l|l|l|}
    \hline
        ~ & Lee et al.\cite{lee:Pill-ID:2012} & NIH Pillbox & NIH Pillbox & CURE\cite{ling:few-shot:2020} & Wong et al.\cite{wong:finegrain:2017} & Tan et al.\cite{tan:comparison:2021} \\ 
        ~ & ~ & (Reference)\cite{yaniv:nlm:2016} & (Consumer)\cite{yaniv:nlm:2016} & ~ & ~ & ~ \\
        \hline
        Total Images & 15,031 & 4,392 & 133,774 & 8,973 & 5,284 & 5,131 \\ \hline
        Labels & None & None & None & Partial Instance & None & Bounding Box \\
        ~ & ~ & ~ & ~ & Segmentation & ~ & ~ \\ \hline
        Publicly Available & No & Yes & No & Yes & No & No \\ \hline
        Synthetic images & No & No & No & Yes & No & No \\ \hline
    \end{tabular}
    \caption{\label{tab:existing_dataset_comparison}Comparison between the different pill recognition datasets.}
\end{table}

\begin{table}[ht]
    \centering
    \begin{tabular}{|l|l|l|l|l|}
    \hline
        ~ & ~ & ~ & MEDISEG & MEDISEG \\
        ~ & NIH Pillbox\cite{yaniv:nlm:2016} & CURE\cite{ling:few-shot:2020} & (3-Pills) (Ours) & (32-Pills) (Ours) \\ \hline
        Number of Pill Images & 4392 & 8973 & 2333 & 8262 \\ \hline
        Pill Categories & 9100 & 196 & 3 & 32 \\ \hline
        Instances per Image & 1 & 1 & 1-6 & 1-11 \\ \hline
        Instances per Category & 1-2 & 18-60 & 1181-1900 & 61-1900 \\ \hline
        Segmentation Labels & None & Partially labelled & All & All \\ \hline
        Synthetic Images & None & Reference images & None & None \\ \hline
    \end{tabular}
    \caption{\label{tab:MEDISEG_dataset_comparison}Comparison of the publicly available datasets and MEDISEG datasets. }
\end{table}

\begin{table}[ht]
    \centering
    \begin{tabular}{|l|c|c|c|c|c|}
    \hline
         & \textbf{fg\_cls\_accuracy} & \textbf{false\_negative} & \textbf{loss\_cls} & \textbf{loss\_rpn\_cls} & \textbf{total\_loss} \\ \hline
        CURE 1-shot & 0.131 & 0.816 & 0.351 & 0.863 & 1.326 \\ \hline
        CURE 5-shot & 0.372 & 0.465 & 0.421 & 0.224 & 0.844 \\ \hline
        CURE 10-shot & 0.558 & 0.342 & 0.320 & 0.133 & 0.674 \\ \hline
        MEDISEG 1-shot & 0.406 & 0.513 & 0.383 & 0.312 & 0.963 \\ \hline
        MEDISEG 5-shot & 0.625 & 0.246 & 0.279 & 0.182 & 0.680 \\ \hline
        MEDISEG 10-shot & 0.740 & 0.210 & 0.191 & 0.059 & 0.445 \\ \hline
    \end{tabular}
    \caption{\label{tab:overlap}Few-shot detection results on the overlap-only test set for models base-trained on CURE and MEDISEG under 1-, 5-, and 10-shot settings.}
\end{table}

\begin{table}[ht]
    \centering
    \begin{tabular}{|l|l|l|l|l|l|l|l|l|l|}
    \hline
        {\textbf{epoch}} & \textbf{train} & \textbf{train} & \textbf{val} & \textbf{val} & {\textbf{precision}} & {\textbf{recall}} & \textbf{F1-} & \textbf{mAP} & \textbf{mAP} \\
        ~ & \textbf{box loss} & \textbf{cls loss} & \textbf{box loss} & \textbf{cls loss} & ~ & ~ & \textbf{Score} & \textbf{50} & \textbf{50-95} \\ \hline
        1 & 0.4425 & 2.3809 & 0.38846 & 1.2704 & 0.98261 & 0.9841 & 0.98335 & 0.99245 & 0.90272 \\ \hline
        2 & 0.37488 & 0.89534 & 0.3755 & 0.83065 & 0.99558 & 0.99392 & 0.99475 & 0.99338 & 0.90903 \\ \hline
        3 & 0.36016 & 0.78955 & 0.37174 & 0.72901 & 0.99494 & 0.99544 & 0.99519 & 0.9933 & 0.91417 \\ \hline
        4 & 0.33755 & 0.68899 & 0.33344 & 0.66842 & 0.99431 & 0.99376 & 0.99403 & 0.99405 & 0.92862 \\ \hline
        5 & 0.31921 & 0.62769 & 0.35102 & 0.60851 & 0.99713 & 0.99555 & 0.99634 & 0.99438 & 0.92478 \\ \hline
        6 & 0.29169 & 0.55505 & 0.33288 & 0.547 & 0.99701 & 0.9974 & 0.9972 & 0.99468 & 0.93055 \\ \hline
        7 & 0.28688 & 0.52533 & 0.32323 & 0.51523 & 0.99719 & 0.99837 & 0.99778 & 0.99436 & 0.93368 \\ \hline
        8 & 0.26482 & 0.48432 & 0.32304 & 0.52219 & 0.9964 & 0.99726 & 0.99683 & 0.99468 & 0.93419 \\ \hline
        9 & 0.25122 & 0.45794 & 0.32895 & 0.49528 & 0.99803 & 0.99656 & 0.99729 & 0.99399 & 0.93168 \\ \hline
        10 & 0.23937 & 0.44254 & 0.3156 & 0.48652 & 0.99436 & 0.99758 & 0.99597 & 0.99356 & 0.93854 \\ \hline
        11 & 0.2252 & 0.41531 & 0.30801 & 0.46587 & 0.99712 & 0.99636 & 0.99674 & 0.9942 & 0.94435 \\ \hline
        12 & 0.21874 & 0.39188 & 0.30673 & 0.45331 & 0.99521 & 0.996 & 0.9956 & 0.9945 & 0.94532 \\ \hline
        13 & 0.20309 & 0.36881 & 0.30307 & 0.43459 & 0.99671 & 0.99714 & 0.99692 & 0.9944 & 0.94943 \\ \hline
        14 & 0.18912 & 0.3489 & 0.30416 & 0.4292 & 0.9977 & 0.99681 & 0.99725 & 0.99355 & 0.94634 \\ \hline
        15 & 0.17162 & 0.32145 & 0.30383 & 0.42545 & 0.99645 & 0.99731 & 0.99688 & 0.9937 & 0.94556 \\ \hline
        16 & 0.16442 & 0.3096 & 0.3001 & 0.41216 & 0.99737 & 0.99776 & 0.99756 & 0.99423 & 0.94911 \\ \hline
        17 & 0.15142 & 0.29073 & 0.29986 & 0.41131 & 0.99679 & 0.997 & 0.99689 & 0.99449 & 0.95049 \\ \hline
        18 & 0.14046 & 0.27143 & 0.29692 & 0.40346 & 0.99711 & 0.99709 & 0.9971 & 0.99416 & 0.95213 \\ \hline
        19 & 0.12508 & 0.24812 & 0.29537 & 0.40358 & 0.9967 & 0.99741 & 0.99705 & 0.9944 & 0.95104 \\ \hline
        20 & 0.11422 & 0.23487 & 0.29379 & 0.40301 & 0.99678 & 0.99695 & 0.99686 & 0.99432 & 0.95002 \\ \hline
    \end{tabular}
    \caption{\label{tab:YOLOv8_3}YOLOv8 training results on MEDISEG (3-Pills)}
\end{table}

\begin{table}[ht]
    \centering
    \begin{tabular}{|l|l|l|l|l|l|l|l|l|l|}
    \hline
        {\textbf{epoch}} & \textbf{train} & \textbf{train} & \textbf{val} & \textbf{val} & {\textbf{precision}} & {\textbf{recall}} & \textbf{F1-} & \textbf{mAP} & \textbf{mAP} \\
        ~ & \textbf{box loss} & \textbf{cls loss} & \textbf{box loss} & \textbf{cls loss} & ~ & ~ & \textbf{Score} & \textbf{50} & \textbf{50-95} \\ \hline
        1 & 0.41587 & 1.4857 & 0.51442 & 1.1569 & 0.35119 & 0.88681 & 0.50313 & 0.38442 & 0.34005 \\ \hline
        2 & 0.40709 & 0.67327 & 0.51442 & 1.1569 & 0.35119 & 0.88681 & 0.50313 & 0.38442 & 0.34005 \\ \hline
        3 & 0.4067 & 0.67247 & 4.007 & 1023.8 & 0.00065 & 0.10815 & 0.00129 & 0.00063 & 0.00019 \\ \hline
        4 & 0.38313 & 0.58268 & 0.50564 & 1.1569 & 0.94721 & 0.88833 & 0.91683 & 0.93286 & 0.81646 \\ \hline
        5 & 0.36867 & 0.50864 & 0.35888 & 0.56471 & 0.9764 & 0.98175 & 0.97907 & 0.99162 & 0.92605 \\ \hline
        6 & 0.33467 & 0.4766 & 0.37976 & 0.55319 & 0.99061 & 0.98163 & 0.9861 & 0.99325 & 0.91393 \\ \hline
        7 & 0.32526 & 0.4561 & 0.34613 & 0.45612 & 0.9955 & 0.994 & 0.99475 & 0.99433 & 0.93544 \\ \hline
        8 & 0.31929 & 0.43968 & 0.33695 & 0.45977 & 0.99331 & 0.99503 & 0.99417 & 0.99416 & 0.94015 \\ \hline
        9 & 0.30966 & 0.41822 & 0.33513 & 0.46915 & 0.99183 & 0.9915 & 0.99166 & 0.99458 & 0.9302 \\ \hline
        10 & 0.30985 & 0.42252 & 0.919 & 3.3255 & 0.78279 & 0.5511 & 0.64682 & 0.63774 & 0.53593 \\ \hline
        11 & 0.29099 & 0.39097 & 0.31898 & 0.43382 & 0.99063 & 0.99396 & 0.99229 & 0.99479 & 0.94887 \\ \hline
        12 & 0.29241 & 0.38455 & 0.31544 & 0.41275 & 0.99207 & 0.99464 & 0.99335 & 0.99439 & 0.94386 \\ \hline
        13 & 0.28295 & 0.38133 & 0.30852 & 0.38144 & 0.99762 & 0.99462 & 0.99612 & 0.99482 & 0.95343 \\ \hline
        14 & 0.2791 & 0.34504 & 0.32661 & 0.39229 & 0.99638 & 0.99459 & 0.99548 & 0.99459 & 0.93819 \\ \hline
        15 & 0.26231 & 0.32706 & 0.29489 & 0.37145 & 0.99421 & 0.99618 & 0.99519 & 0.99476 & 0.95859 \\ \hline
        16 & 0.2489 & 0.31283 & 0.29645 & 0.34421 & 0.99575 & 0.99593 & 0.99584 & 0.99448 & 0.95807 \\ \hline
        17 & 0.24831 & 0.31448 & 0.29125 & 0.34961 & 0.99587 & 0.99606 & 0.99596 & 0.99459 & 0.96038 \\ \hline
        18 & 0.23418 & 0.29033 & 0.28761 & 0.32829 & 0.99577 & 0.99766 & 0.99671 & 0.99459 & 0.96296 \\ \hline
        19 & 0.22426 & 0.27393 & 0.28045 & 0.31022 & 0.99818 & 0.99794 & 0.99806 & 0.99452 & 0.96419 \\ \hline
        20 & 0.21401 & 0.26267 & 0.27962 & 0.30339 & 0.99611 & 0.9982 & 0.99715 & 0.99462 & 0.96473 \\ \hline
    \end{tabular}
    \caption{\label{tab:YOLOv9_3}YOLOv9 training results on MEDISEG (3-Pills)}
\end{table}

\begin{table}[ht]
    \centering
    \begin{tabular}{|l|l|l|l|l|l|l|l|l|l|}
    \hline
        {\textbf{epoch}} & \textbf{train} & \textbf{train} & \textbf{val} & \textbf{val} & {\textbf{precision}} & {\textbf{recall}} & \textbf{F1-} & \textbf{mAP} & \textbf{mAP} \\
        ~ & \textbf{box loss} & \textbf{cls loss} & \textbf{box loss} & \textbf{cls loss} & ~ & ~ & \textbf{Score} & \textbf{50} & \textbf{50-95} \\ \hline
        1 & 0.73515 & 5.5465 & 0.79503 & 3.4362 & 0.43798 & 0.24101 & 0.31093 & 0.18367 & 0.14918 \\ \hline
        2 & 0.604 & 2.6415 & 0.68777 & 2.5692 & 0.5086 & 0.34743 & 0.41284 & 0.31229 & 0.25544 \\ \hline
        3 & 0.53917 & 2.0917 & 0.67365 & 2.1735 & 0.55653 & 0.41183 & 0.47337 & 0.41752 & 0.33367 \\ \hline
        4 & 0.51319 & 1.7586 & 0.63762 & 1.9514 & 0.59987 & 0.47221 & 0.52844 & 0.50077 & 0.40223 \\ \hline
        5 & 0.49068 & 1.4985 & 0.62279 & 1.8043 & 0.60842 & 0.49662 & 0.54686 & 0.54505 & 0.43708 \\ \hline
        6 & 0.4588 & 1.3055 & 0.58566 & 1.7361 & 0.56962 & 0.54908 & 0.55916 & 0.54817 & 0.45021 \\ \hline
        7 & 0.43396 & 1.1668 & 0.57897 & 1.6617 & 0.603 & 0.54142 & 0.57055 & 0.56817 & 0.46201 \\ \hline
        8 & 0.41625 & 1.0688 & 0.58633 & 1.596 & 0.58393 & 0.56595 & 0.5748 & 0.57867 & 0.4733 \\ \hline
        9 & 0.39766 & 0.97933 & 0.58376 & 1.6342 & 0.58132 & 0.55538 & 0.56805 & 0.60054 & 0.48778 \\ \hline
        10 & 0.38401 & 0.90746 & 0.57247 & 1.5779 & 0.60574 & 0.56414 & 0.5842 & 0.60726 & 0.49805 \\ \hline
        11 & 0.3661 & 0.84459 & 0.56435 & 1.6035 & 0.61865 & 0.55074 & 0.58272 & 0.60565 & 0.49435 \\ \hline
        12 & 0.34908 & 0.79938 & 0.56286 & 1.5538 & 0.68249 & 0.54644 & 0.60693 & 0.62448 & 0.51006 \\ \hline
        13 & 0.33154 & 0.75038 & 0.55853 & 1.5104 & 0.58957 & 0.58816 & 0.58886 & 0.62737 & 0.51686 \\ \hline
        14 & 0.31673 & 0.71364 & 0.55565 & 1.5752 & 0.68815 & 0.56191 & 0.61866 & 0.62418 & 0.51367 \\ \hline
        15 & 0.30624 & 0.68362 & 0.55669 & 1.5129 & 0.66536 & 0.57576 & 0.61733 & 0.62655 & 0.51197 \\ \hline
        16 & 0.29127 & 0.65046 & 0.55247 & 1.5134 & 0.64348 & 0.56754 & 0.60313 & 0.62241 & 0.51141 \\ \hline
        17 & 0.27615 & 0.62352 & 0.54751 & 1.4935 & 0.60451 & 0.58423 & 0.5942 & 0.62307 & 0.51277 \\ \hline
        18 & 0.2628 & 0.59748 & 0.55078 & 1.5027 & 0.6278 & 0.57369 & 0.59953 & 0.6216 & 0.50946 \\ \hline
    \end{tabular}
    \caption{\label{tab:YOLOv8_32}YOLOv8 training results on MEDISEG (32-Pills)}
\end{table}

\begin{table}[ht]
    \centering
    \begin{tabular}{|l|l|l|l|l|l|l|l|l|l|}
    \hline
        {\textbf{epoch}} & \textbf{train} & \textbf{train} & \textbf{val} & \textbf{val} & {\textbf{precision}} & {\textbf{recall}} & \textbf{F1-} & \textbf{mAP} & \textbf{mAP} \\
        ~ & \textbf{box loss} & \textbf{cls loss} & \textbf{box loss} & \textbf{cls loss} & ~ & ~ & \textbf{Score} & \textbf{50} & \textbf{50-95} \\ \hline
        1 & 0.50243 & 2.9938 & 0.62794 & 2.4668 & 0.56949 & 0.47 & 0.51498 & 0.4475 & 0.36437 \\ \hline
        2 & 0.49417 & 1.6132 & 0.61723 & 2.0512 & 0.19794 & 0.44166 & 0.27337 & 0.18848 & 0.14914 \\ \hline
        3 & 0.48833 & 1.3587 & 0.59486 & 1.5773 & 0.51456 & 0.62207 & 0.56323 & 0.60872 & 0.48423 \\ \hline
        4 & 0.46465 & 1.1243 & 0.5547 & 1.4804 & 0.71965 & 0.55804 & 0.62862 & 0.62773 & 0.51653 \\ \hline
        5 & 0.44736 & 0.97132 & 0.54143 & 1.3826 & 0.70975 & 0.64348 & 0.67499 & 0.68883 & 0.56494 \\ \hline
        6 & 0.42239 & 0.84014 & 0.52069 & 1.2929 & 0.70095 & 0.62673 & 0.66177 & 0.6969 & 0.5739 \\ \hline
        7 & 0.39895 & 0.72349 & 0.51985 & 1.2 & 0.7608 & 0.62878 & 0.68852 & 0.72059 & 0.59146 \\ \hline
        8 & 0.38331 & 0.63141 & 0.50532 & 1.1008 & 0.7266 & 0.70616 & 0.71623 & 0.74005 & 0.60873 \\ \hline
        9 & 0.3748 & 0.58613 & 0.49972 & 1.0999 & 0.80646 & 0.67822 & 0.7368 & 0.7695 & 0.63651 \\ \hline
        10 & 0.3555 & 0.53785 & 0.47401 & 0.94367 & 0.7703 & 0.71289 & 0.74048 & 0.77923 & 0.65513 \\ \hline
        11 & 0.33482 & 0.47364 & 0.48329 & 0.97246 & 0.77822 & 0.71156 & 0.7434 & 0.78164 & 0.65551 \\ \hline
        12 & 0.32306 & 0.45573 & 0.47604 & 0.97515 & 0.77924 & 0.72021 & 0.74856 & 0.77782 & 0.65508 \\ \hline
        13 & 0.30168 & 0.42396 & 0.47352 & 1.0052 & 0.82042 & 0.67782 & 0.74233 & 0.77556 & 0.64838 \\ \hline
        14 & 0.28736 & 0.38669 & 0.46723 & 0.97432 & 0.82982 & 0.70393 & 0.76171 & 0.78655 & 0.66446 \\ \hline
        15 & 0.26959 & 0.36723 & 0.45504 & 0.93764 & 0.80321 & 0.73529 & 0.76775 & 0.79458 & 0.67314 \\ \hline
        16 & 0.25225 & 0.34321 & 0.45493 & 0.92608 & 0.83184 & 0.7097 & 0.76593 & 0.80105 & 0.68232 \\ \hline
        17 & 0.23883 & 0.32973 & 0.45395 & 0.92773 & 0.81874 & 0.71622 & 0.76406 & 0.78599 & 0.67112 \\ \hline
        18 & 0.22576 & 0.31198 & 0.44453 & 0.90586 & 0.80074 & 0.73438 & 0.76613 & 0.79795 & 0.68051 \\ \hline
        19 & 0.20858 & 0.29412 & 0.44598 & 0.89233 & 0.80779 & 0.73527 & 0.76983 & 0.79839 & 0.68159 \\ \hline
        20 & 0.19251 & 0.27755 & 0.4415 & 0.88509 & 0.81248 & 0.73708 & 0.77295 & 0.80079 & 0.68352 \\ \hline
    \end{tabular}
    \caption{\label{tab:YOLOv9_32}YOLOv9 training results on MEDISEG (32-Pills)}
\end{table}

\end{document}